\documentclass{article}

\PassOptionsToPackage{numbers, compress}{natbib}
\usepackage[preprint]{neurips_2021}

\usepackage[utf8]{inputenc} 
\usepackage[T1]{fontenc}    
\usepackage{hyperref}      
\usepackage{url}         
\usepackage{booktabs}      
\usepackage{amsfonts}    
\usepackage{nicefrac}     
\usepackage{microtype}   
\usepackage{xcolor}   
\usepackage{adjustbox}
\usepackage{graphicx}

\usepackage{bm}
\usepackage{epstopdf}
\usepackage{amsmath}
\usepackage{amsfonts}
\usepackage{balance}
\usepackage{multirow}
\usepackage{color}
\usepackage{fontenc} 
\usepackage{amsmath}
\usepackage{caption}
\usepackage{subcaption}
\usepackage{lineno}
\usepackage[normalem]{ulem}
\usepackage{booktabs}
\usepackage{comment}
\usepackage{mathtools}
\usepackage{algorithm,algpseudocode}
\usepackage{stfloats}
\usepackage{amsthm}
\usepackage{mathtools}
\usepackage{amsfonts,amssymb}
\usepackage{enumitem}

\useunder{\uline}{\ul}{}

\newcommand{\model}{\textbf{OOD-GNN} }
\newcommand{\mymodel}{\textbf{OOD-GNN}}
\newcommand{\indep}{\perp \!\!\! \perp}
\theoremstyle{definition}

\theoremstyle{remark}

\theoremstyle{proposition}
\newtheorem{prop}{Proposition}
\newcommand{\defeq}{\vcentcolon=}

\title{OOD-GNN: Out-of-Distribution Generalized \\ Graph Neural Network}

\author{
Haoyang Li,\;
Xin Wang,\;
Ziwei Zhang,\;
Wenwu Zhu\\
Tsinghua University\\
\small \texttt{lihy18@mails.tsinghua.edu.cn},\;
\small \texttt{\{xin\_wang, zwzhang, wwzhu\}@tsinghua.edu.cn}\\
}

\begin{document}

\maketitle

\begin{abstract}
Graph neural networks (GNNs) have achieved impressive performance when testing and training graph data come from identical distribution.
However, existing GNNs lack out-of-distribution generalization abilities so that their performance substantially degrades when there exist distribution shifts between testing and training graph data.
To solve this problem, in this work, we propose an out-of-distribution generalized graph neural network (\mymodel) for achieving satisfactory performance on unseen testing graphs that have different distributions with training graphs.
Our proposed \model employs a novel nonlinear graph representation decorrelation method utilizing random Fourier features, which encourages the model to eliminate the statistical dependence between relevant and irrelevant graph representations through iteratively optimizing the sample graph weights and graph encoder. 
We further present a global weight estimator to learn weights for training graphs such that variables in graph representations are forced to be independent.
The learned weights help the graph encoder to get rid of spurious correlations and, in turn, concentrate more on the true connection between learned discriminative graph representations and their ground-truth labels.
We conduct extensive experiments to validate the out-of-distribution generalization abilities on two synthetic and 12 real-world datasets with distribution shifts.
The results demonstrate that our proposed \model significantly outperforms state-of-the-art baselines.
\end{abstract}

\section{Introduction}
\label{sec:introduction}

Graph structured data is ubiquitous in the real world, e.g., biology networks~\cite{barabasi2004network}, social networks~\cite{easley2010networks}, molecular graphs~\cite{wu2018moleculenet}, knowledge graphs~\cite{wang2017knowledge}, etc. 
Recently, deep learning models on graphs, especially graph neural networks (GNNs)~\cite{kipf2016semi, velivckovic2017graph, xu2018powerful}, have increasingly emerged as prominent approaches for representation learning of graphs~\cite{hamilton2020graph}.
Significant methodological advances have been made in the field of GNNs, which have achieved promising performance in a wide variety of applications~\cite{li2017learning, fan2020graph, li2021intention, zitnik2018modeling}. 

Despite their enormous success, the existing GNN approaches for graph representation learning generally assume that the testing and training graph data are independently sampled from the identical distribution, i.e., the I.I.D. assumption. 
In many real-world scenarios, however, it is difficult to guarantee this assumption to be valid. 
In particular, the testing distribution may suffer unobserved or uncontrolled shifts compared with the training distribution.  
For example, in the field of drug discovery, the prediction of biochemical properties of molecules is commonly trained on limited available experimental data, but the model needs to be tested on an extraordinarily diverse and combinatorially large universe of candidate molecules~\cite{sterling2015zinc, bohacek1996art}. 
The model performance of existing methods can be substantially degraded under distribution shifts due to the lack of out-of-distribution (\textbf{OOD}) generalization ability in realistic data splits~\cite{hu2020open, wu2018moleculenet}. 
Therefore, it is of paramount importance to learn GNNs capable of out-of-distribution generalization and achieve relatively stable performances under distribution shifts, especially for some high-stake applications, e.g., medical diagnosis~\cite{li2020graph}, criminal justice~\cite{han2020risk}, financial analysis~\cite{yang2019using}, and molecular prediction~\cite{wu2018moleculenet}, etc.

\begin{figure*}[t]
\centering
\begin{subfigure}{.42\linewidth}

  \centering
  \includegraphics[width=.99\linewidth]{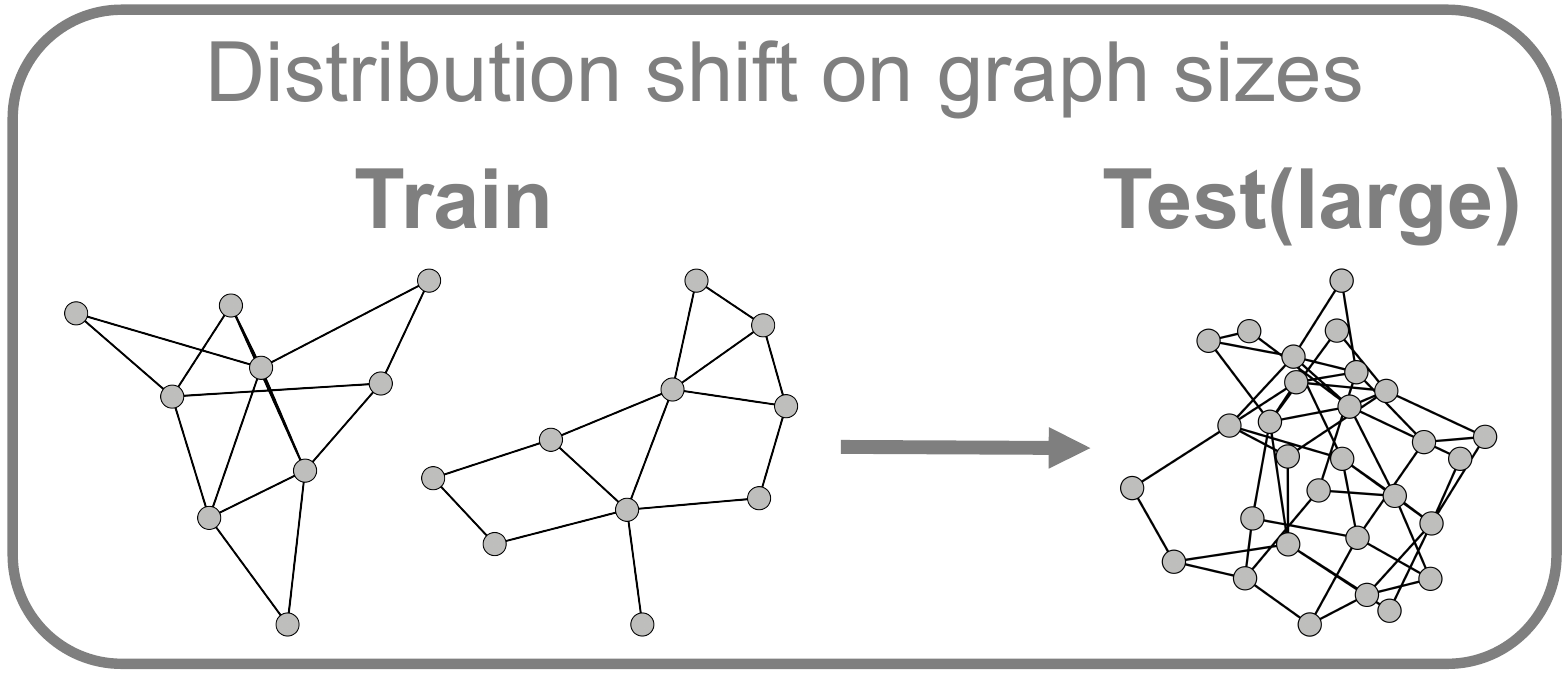} 
  \caption{TRIANGLES}
  \label{fig:example_triangles}
\end{subfigure}
\begin{subfigure}{.42\linewidth}
  \centering
  \includegraphics[width=.99\linewidth]{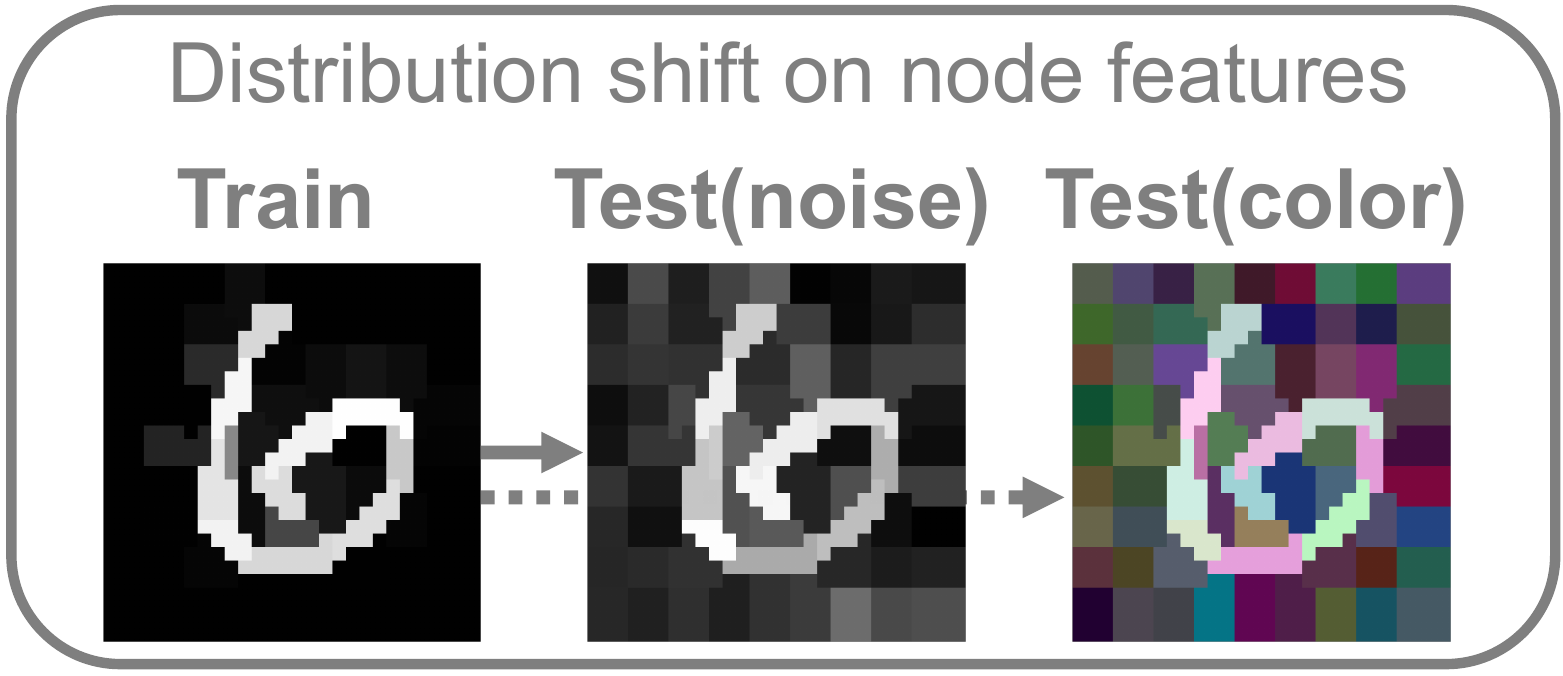} 
  \caption{MNIST-75SP: Super-pixel Graphs}
  \label{fig:example_mnist}
\end{subfigure}

\begin{subfigure}{.84\linewidth}
  \centering
  \includegraphics[width=.99\linewidth]{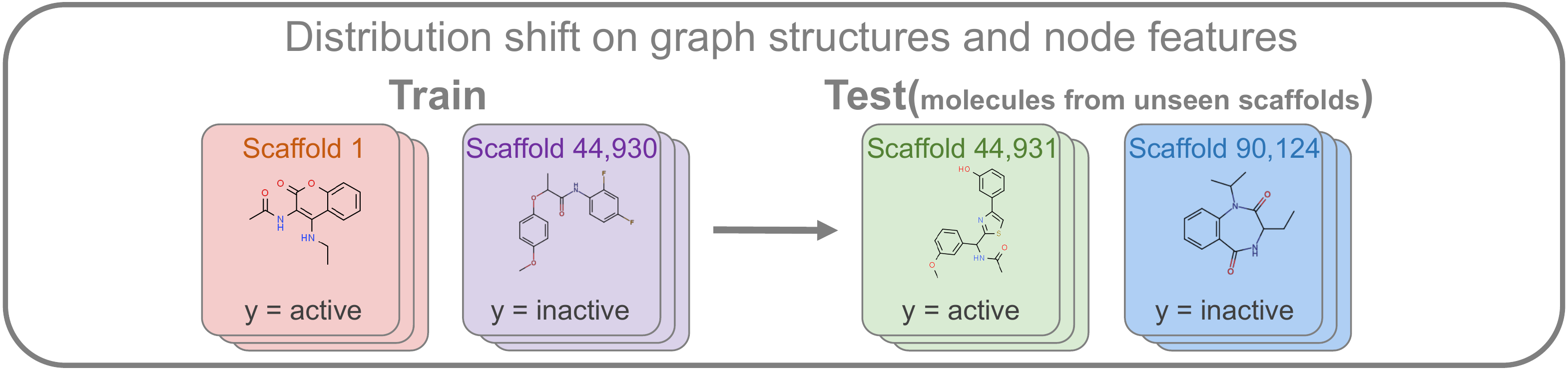} 
  \caption{OGB Molecule Dataset~\cite{koh2020wilds}. For validating OOD generalization, this dataset is split based on the scaffolds (i.e., two-dimensional structural frameworks) of molecules. The testing set consists of structurally distinct molecules with scaffolds that are not in the training set.}
  \label{fig:example_ogb}
\end{subfigure}
\caption{
Examples of out-of-distribution testing graphs under complex distribution shifts. Figure~\ref{fig:example_triangles} denotes the models are trained on small graphs but tested on larger graphs. Figure~\ref{fig:example_mnist} denotes the models trained with clean node features but tested with noisy features. Figure~\ref{fig:example_ogb} represents a more realistic and challenging case, i.e., distribution shifts exist on both graph structures and node features. 
}
\label{figure:example_synthetic}
\end{figure*}

Some pioneering works~\cite{knyazev2019understanding, yehudai2020size, bevilacqua2021size} focus on the size generalization problem by testing on larger graphs than the training graphs.
Besides size generalization, the capability of out-of-distribution generalization for GNNs is not explored until recently~\cite{xu2020neural}.
In out-of-distribution scenarios, when there exist complex heterogeneous distribution shifts, the performance of current GNN models can degrade substantially, which is mainly induced by the spurious correlations. 
The spurious correlations intrinsically come from the subtle correlations between irrelevant representations and relevant representations~\cite{arjovsky2019invariant,tu2020empirical}. 
For example, in the field of drug discovery (see Figure~\ref{fig:example_ogb}), the GNN models trained on molecules with one group of scaffolds (two-dimensional structural frameworks of molecules) may learn the spurious correlations between the scaffolds and labels (i.e., whether some drug can inhibit HIV replication)~\cite{wu2018moleculenet, hu2020open}. 
When tested on molecules with different scaffolds (out-of-distribution testing molecules), the existing GNN models may make incorrect predictions based on the spurious correlations.

In this paper, we propose to learn decorrelated graph representations through sample reweighting~\cite{ren2018learning, kuang2018stable} to eliminate the dependence between irrelevant and relevant representations, which is one of the major causes of degrading model performance under distribution shifts.
However, learning decorrelated graph representations to improve out-of-distribution generalization for GNNs is fundamentally different from traditional methods and thus remains largely unexplored and challenging.
Specifically, it poses the following challenges. 
\begin{itemize}[leftmargin = 0.5cm]
    \item GNNs fuse heterogeneous information from node features and graph structures such that the complex and unobserved non-linear dependencies among representations are much more difficult to be measured and eliminated than the linear cases for decorrelation of non-graph data.
    \item Although sample reweighting is effective on small datasets, for real-world large-scale graphs, it is inefficient or even infeasible to consistently learn a global weight for each graph in the dataset due to the high computational complexity and excessive storage consumption.
\end{itemize}

To tackle these challenges, we propose a novel out-of-distribution generalized graph neural network (\mymodel) capable of handling graph distribution shifts in complex and heterogeneous situations.
In particular, we first propose to eliminate the statistical dependence between relevant and irrelevant graph representations of the graph encoder by a novel nonlinear graph representation decorrelation method utilizing random Fourier features~\cite{rahimi2007random, zhang2021deep, li2019towards}, which scales linearly with the sample size and can get rid of unexpected spurious correlations.
Next, to reduce computational complexity, we present a scalable global-local weight estimator to learn the sample weight for each graph.
The local weights for a mini-batch of graphs and global weights for the entire graphs are optimized jointly to effectively maintain the consistency of weights over the whole graph dataset~\cite{zhang2021deep,wu2018unsupervised, he2020momentum}. 
Finally, the parameters of the graph encoder and sample weights for graph representation decorrelation are optimized iteratively to learn discriminant graph representations for predictions.

We conduct extensive experiments on both synthetic graph datasets and well-known real-world graph benchmarks. 
The experimental results demonstrate that the representations learned from \model can achieve substantial performance gains on the graph prediction tasks, including graph classification and regression, under distribution shifts.

The contributions of this paper are summarized as follows:
\begin{itemize}[leftmargin = 0.5cm]
	\item We propose a novel out-of-distribution generalized graph neural network (\mymodel) capable of learning out-of-distribution (OOD) generalized graph representation under complex distribution shifts.
	\item We propose a nonlinear graph representation decorrelation method based on random Fourier features and sample reweighting. The decorrelated graph representations can substantially improve the out-of-distribution generalization ability in various OOD graph prediction benchmarks.
	\item We present a scalable global-local weight estimator to learn graph weights for the whole dataset consistently and efficiently. Extensive empirical results show that \model greatly outperforms baselines on various graph prediction benchmarks under distribution shifts.
\end{itemize}

We review related works in Section \ref{section:relatedwork}.
In Section \ref{section:method}, we describe the problem formulation and the details of our proposed \mymodel. 
Section \ref{section:experiments} presents the experimental results including quantitative comparisons on both synthetic and real-world datasets, ablation studies, complexity analysis, hyper-parameter sensitivity, etc. 
Finally, we conclude our work in Section \ref{section:conclusions}.

\section{Related Works}
\label{section:relatedwork}
\textbf{Graph Neural Network.}
GNNs~\cite{kipf2016semi, velivckovic2017graph, xu2018powerful} have been attracting considerable attention in recent years because of their notable success in representing graph-structure data. 
They generally utilize a message-passing paradigm, which combines node features and graph topology to update node embeddings.
To obtain the representation of the entire graph, graph pooling~\cite{xu2018powerful, lee2019self, zhang2021hierarchical} is adopted to summarize node embeddings.
Many GNNs and their variants~\cite{hamilton2017inductive, morris2019weisfeiler, yu2021self, li2021higher} have been proposed, achieving state-of-the-art performance on various graph tasks, including node classification~\cite{kipf2016semi},  link prediction~\cite{schlichtkrull2018modeling}, and graph classification~\cite{xu2018powerful, gao2021higher}.
Despite their successes, the performance of GNNs drops substantially when there are distribution shifts between training and testing graphs~\cite{koh2020wilds, wu2018moleculenet, hu2020open}.
The existing works largely ignore the out-of-distribution generalization ability of GNNs, which is crucial to realistic applications deployed in the wild~\cite{koh2020wilds}.

\textbf{Size generalization of GNNs.}
The main goal of size generalization is to make GNNs work well on testing graphs whose size distribution is different from that of training graphs~\cite{santoro2018measuring, knyazev2019understanding, yehudai2020size, saxton2019analysing, velivckovic2019neural}. In these works, GNNs are usually trained on relatively small graphs and then generalize to larger graphs with the help of attention mechanisms~\cite{knyazev2019understanding} and self-supervised learning~\cite{yehudai2020size}. 
\citet{bevilacqua2021size} propose to learn invariant graph representations for extrapolations with a predefined structural causal model.
However, most existing methods only test on graphs of different sizes and ignore more realistic and challenging settings where the distribution shifts emerge in the graph topologies and node features.

\textbf{The expressiveness of GNNs.}
The Weisfeiler-Lehman graph isomorphism test is most commonly used to measure the expressiveness power of GNNs~\cite{xu2018powerful, morris2019weisfeiler}.
Assuming appropriate optimization, a more expressive GNN can achieve smaller error on the training data~\cite{loukas2020graph}.
Some works~\cite{garg2020generalization, verma2019stability} also study the generalization capability of GNNs over the training distribution.
These works are orthogonal to out-of-distribution generalization, including unseen graph topological structures and features studied in this paper.
The findings in~\cite{xu2020neural} show that encoding task-specific non-linearities in the GNN architecture or features can improve the out-of-distribution generalization. 
However, it is largely unknown in practice that how to enhance the generalization ability of GNNs when there are distribution shifts between training and testing graphs.

\textbf{Representation decorrelation.}
The spurious correlation between the irrelevant (non-critical) representations and labels is recognized as one major cause of model degradation under distribution shifts~\cite{cogswell2015reducing, gu2018regularizing, arpit2019predicting, song2021influence}.
Some pioneering works adopt regularizers to penalize high correlation explicitly~\cite{hebiri2012correlations, rodriguez2016regularizing, cogswell2015reducing}. 
However, these methods could introduce a substantial computational overhead, yield marginal improvements, or require extra supervision to control the strength of the penalty.
There are also some works learning decorrelated representations with sample reweighting~\cite{kuang2018stable, shen2020stable, kuang2020stable, zhang2018removing}, which is shown effective in improving the generalization ability theoretically (e.g., SRDO~\cite{shen2020stable}) and empirically (e.g., DWR~\cite{kuang2020stable}).
However, most of these methods are proposed under linear settings. 
In contrast, GNNs fuse heterogeneous information from node features and graph topological structures so that there exist complex and unobserved non-linear dependencies among representations.
The linear sample reweighting methods can not be applied to eliminate non-linear dependencies for the decorrelation of graph data. 
We also observe a significant performance drop in the experiments if only linear dependencies between representations are eliminated.
The effectiveness of non-linear decorrelation methods (e.g, ReBias~\cite{bahng2020learning}, StableNet~\cite{zhang2021deep}) is validated on images recently. 
However, non-linear decorrelation on graphs remains largely unexplored.

\textbf{Disentangled graph representation learning.}
Disentangled representation learning has gained considerable attention in the last few years, aiming to characterize the various underlying explanatory factors behind the observed data in different parts of the factorized vector representation \cite{bengio2013representation}. 
The existing efforts about disentangled representation learning are originally designed for computer vision \cite{hsieh2018learning, ma2018disentangled}.
More recently, some efforts generalizing disentangled representation learning for graph data have been proposed. 
For example, some works utilize the dynamic routing mechanism to disentangle latent factors for node-level representation learning (e.g., DisenGCN~\cite{ma2019disentangled}) and for the whole graph representation learning~\cite{yang2020factorizable}. 
FactorGCN~\cite{yang2020factorizable} decomposes the input graph into several interpretable factor graphs for graph-level disentangled representations, which is a state-of-the-art disentangled GNN model.
However, these methods force the representations to be factorized vectors and can change the semantic implication of the graph representations. 
In addition, their results on downstream tasks may be degraded due to the trade-off dilemma \cite{burgess2018understanding} between disentanglement and performance. 
We observe in the experiments that the SOTA graph disentanglement method FactorGCN fails to achieve promising results under complex distribution shifts.
In contrast, for achieving out-of-distribution generalization, our method is able to learn graph weights while keeping the semantic implication of the representations unaffected, leading to better OOD generalization. 

\section{Method}
\label{section:method}

\subsection{Notations and Problem Formulation}

Let $\mathbf{G}^{tr} = \{ G_n \}^{N^{tr}}_{n=1}$ and $\mathbf{G}^{te} = \{ G_n \}^{N^{te}}_{n=1}$ be the training and testing graph dataset, which are under distribution shifts, i.e., ${\rm P}(\mathbf{G}^{tr}) \neq {\rm P}(\mathbf{G}^{te})$.
$\mathbf{G}^{te}$ is unobserved in the training stage.
A graph encoder $\Phi: \mathcal{G} \rightarrow \mathcal{Z}$ is a mapping from the input graph space $\mathcal{G}$ to a $d$-dimensional representation space $\mathcal{Z}$. In this work, we consider $\Phi$ as GNNs.
$\mathcal{R}: \mathcal{Z} \rightarrow \mathcal{Y}$ is a classifier, mapping the representation space $\mathcal{Z}$ to the label space $\mathcal{Y}$. 
$\mathbf{G}, \mathbf{Z}, \mathbf{Y}$ denote sets of random variables in  $\mathcal{G}, \mathcal{Z}, \mathcal{Y}$, respectively.
Denote graph representations for $\mathbf{G}^{tr}$ as $\mathbf{Z} \subset \mathbb{R}^{N^{tr} \times d}$. $\mathbf{Z}_{n*}$ denotes the representation of the $n$-th graph and $\mathbf{Z}_{*i}$ is the random variable corresponding to the $i$-th dimension of $\mathbf{Z}$.
Graph weights are $\mathbf{W} = \{ w_n \}^{N^{tr}}_{n=1}$, where $w_n$ is the weight for the $n$-th graph $G_n$ in $\mathbf{G}^{tr}$ and we constrain $\sum_{n=1}^{N^{tr}} w_n = N^{tr}$.
By jointly optimizing the graph encoder $\Phi$, classifier $\mathcal{R}$, and graph weights $\mathbf{W}$, we aim to eliminate the statistical dependence of all dimensions in representation $\mathbf{Z}$ such that the predictor $\mathcal{R} \circ \Phi: \mathcal{G} \rightarrow \mathcal{Y}$ can achieve satisfactory generalization performance when testing on out-of-distribution graphs ${\rm P}(\mathbf{G}^{te})$.

\subsection{Statistical Independence with Graph Reweighting}

The correlation between relevant and irrelevant parts in representations is recognized as the main performance obstacle when ${\rm P}(\mathbf{G}^{tr}) \neq {\rm P}(\mathbf{G}^{te})$, i.e., OOD testing data~\cite{ilse2020diva, kuang2018stable}.
The relevant parts in representations denote the truly discriminant information to predict ground-truth labels, which are invariant under distribution shifts, e.g., the predictive functional blocks of molecules. On the other hand, the irrelevant parts include non-informative features  
that could change across different domains, e.g., scaffold structure in predicting molecule functions.
GNNs fuse available information from node features and graph topologies into a unified low-dimensional representation for each graph.
 So it is difficult or even infeasible to distinguish which dimensionality in the representation denotes relevant and irrelevant parts without extra supervision, which is unavailable and expensive to collect.
Therefore, we propose to encourage the graph encoder to eliminate the statistical dependence of all dimensions in the graph representation.  
Formally, we expect 
\begin{equation}
\mathbf{Z}_{*i} \indep \mathbf{Z}_{*j}, \forall i,j \in [1, d], i \neq j.
\end{equation}

For measuring the independence between continuous random variables $\mathbf{Z}_{*i}$ and $\mathbf{Z}_{*j}$ in $d$-dimensional graph representation space $\mathcal{Z}$, it is inapplicable to resort to histogram-based measures unless $d$ is small enough.
So we introduce Hilbert-Schmidt Independence Criterion (HSIC)~\cite{gretton2005measuring}. Specifically, consider a measurable, positive definite kernel $k_{\mathbf{Z}_{*i}}$ on the domain of random variable $\mathbf{Z}_{*i}$. Denote the corresponding Reproducing Kernel Hilbert Spaces (RKHS) by $\mathcal{H}_{\mathbf{Z}_{*i}}$. 
HSIC is defined as HSIC$(\mathbf{Z}_{*i}, \mathbf{Z}_{*j}) \defeq \lVert C_{\mathbf{Z}_{*i}, \mathbf{Z}_{*j}} \rVert_{\rm HS}^2$, where $C_{\mathbf{Z}_{*i}, \mathbf{Z}_{*j}}$ is the cross-covariance operator in the RKHS of $k_{\mathbf{Z}_{*i}}$ and $k_{\mathbf{Z}_{*j}}$.
The independence can be determined as follows~\cite{fukumizu2007kernel}.
\begin{prop}
	Assume $\mathbb{E} [k_{\mathbf{Z}_{*i}} (\mathbf{Z}_{*i}, \mathbf{Z}_{*i})] \textless \infty$ and $\mathbb{E} [k_{\mathbf{Z}_{*j}} (\mathbf{Z}_{*j}, \mathbf{Z}_{*j})] \textless \infty$, and $k_{\mathbf{Z}_{*i}}k_{\mathbf{Z}_{*j}}$ is a characteristic kernel, then
	\begin{equation}
	{\rm HSIC}(\mathbf{Z}_{*i}, \mathbf{Z}_{*j}) = 0 \Leftrightarrow \mathbf{Z}_{*i} \indep \mathbf{Z}_{*j}.
	\end{equation}
\end{prop}
Although a finite-sample estimate of HSIC has been used in practice for statistical testing~\cite{gretton2005measuring}, it is infeasible to be utilized for training the graph encoder $\Phi$ on large-scale datasets (e.g., the OGBG-MOLHIV dataset in our experiments contains 41,127 graphs). 
The bottleneck lies in that the computational cost of HSIC grows as the batch size of training data increases. 
We therefore consider the squared Frobenius norm $\lVert \widehat{C}_{\mathbf{Z}_{*i}, \mathbf{Z}_{*j}} \rVert_{\rm F}^2$, an analogue corresponding to the HSIC in Euclidean space\footnote{In a finite-dimensional Euclidean space, the Hilbert–Schmidt norm $\lVert \cdot \rVert_{\rm HS}$ is identical to the Frobenius norm.}~\cite{bahng2020learning, zhang2021deep}, where $\widehat{C}_{\mathbf{Z}_{*i}, \mathbf{Z}_{*j}}$ is the partial cross-covariance matrix defined as:
\begin{equation}
\label{equation:partial_cross_covariance}
\begin{array}{r}
\widehat{C}_{\mathbf{Z}_{*i}, \mathbf{Z}_{*j}} = \frac{1}{N^{tr}-1} {\sum_{n=1}^{N^{tr}}} \left[ \left( f(\mathbf{Z}_{ni}) - \frac{1}{N^{tr}}\sum_{m=1}^{N^{tr}} f(\mathbf{Z}_{mi})  \right)^\top  \right. \\
\left. \cdot \left( g(\mathbf{Z}_{nj}) - \frac{1}{N^{tr}}\sum_{m=1}^{N^{tr}} g(\mathbf{Z}_{mj})  \right) \right],
\end{array}
\end{equation}
where $\mathbf{Z}_{ni}$ and $\mathbf{Z}_{nj}$ denote the value of random variables $\mathbf{Z}_{*i}$ and $\mathbf{Z}_{*j}$ given the input graph $G_n$.
\begin{equation}
\label{equation:rff}
\begin{aligned}
	f(\mathbf{Z}_{*i}) &\defeq (f_1(\mathbf{Z}_{*i}), f_2(\mathbf{Z}_{*i}), \dots, f_{Q}(\mathbf{Z}_{*i})), \\	
	g(\mathbf{Z}_{*j}) &\defeq (g_1(\mathbf{Z}_{*j}), g_2(\mathbf{Z}_{*j}), \dots, g_{Q}(\mathbf{Z}_{*j})), 
\end{aligned}
\end{equation}
with $f_q(\mathbf{Z}_{*i}), g_q(\mathbf{Z}_{*j}) \in \mathcal{H}_{\rm RFF}, \forall q \in [1, Q].$
$\mathcal{H}_{\rm RFF} = \{h: x \rightarrow \sqrt{2} {\rm cos}(wx+\phi)|w \sim \mathcal{N}(0,1), \phi \sim {\rm Uniform(0, 2\pi)} \}$ denotes the random Fourier features function space, from which we select $Q$ functions. 
In a nutshell, random Fourier feature (RFF) is an effective technique to approximate kernel-based independence test~\cite{strobl2017approximate, zhang2021deep, li2019towards}.
Note that as $Q$ grows, the accuracy of independence judgement increases. And $Q=5$ is solid enough to measure the independence of random variables in practice~\cite{strobl2017approximate, zhang2021deep}.
 
Using the independence criterion above, we elaborate on graph reweighting which encourages the independence of the variables in graph representation.
Define the graph weights $\mathbf{W} = \{ w_n \}^{N^{tr}}_{n=1}$ where $w_n \in \mathbb{R}$ is the learnable weight for the $n$-th graph $G_n$ in the training set. The graph weights can be directly utilized into Eq.~\eqref{equation:partial_cross_covariance}, so the partial cross-covariance matrix can be calculated as:
\begin{equation}
\label{equation:partial_cross_covariance_weight}
\begin{array}{r}
\widehat{C}_{\mathbf{Z}_{*i}, \mathbf{Z}_{*j}}^{\mathbf{W}} =\frac{1}{N^{tr}-1} {\sum_{n=1}^{N^{tr}}} \left[ \left( w_n f(\mathbf{Z}_{ni}) - \frac{1}{N^{tr}}\sum_{m=1}^{N^{tr}} w_m f(\mathbf{Z}_{mi})  \right)^\top  \right. \\
\left. \cdot \left( w_n g(\mathbf{Z}_{nj}) - \frac{1}{N^{tr}}\sum_{m=1}^{N^{tr}} w_m g(\mathbf{Z}_{mj})  \right) \right].
\end{array}
\end{equation}
The learnable graph weight $\mathbf{W}$ participates in the optimization process to eliminate the dependence between representations to the greatest possible extent by minimizing the squared Frobenius norm of the partial cross-covariance matrix $\lVert \widehat{C}_{\mathbf{Z}_{*i}, \mathbf{Z}_{*j}}^{\mathbf{W}} \rVert_{\rm F}^2$ in Eq.~\eqref{equation:partial_cross_covariance_weight}.

For the optimization, we iteratively optimize the graph weights $\mathbf{W}$, graph encoder $\Phi$, and classifier $\mathcal{R}$:
\begin{align}
	\label{equation:objective1} \Phi^{*},  \mathcal{R}^{*} &= {\rm argmin}_{\Phi,  \mathcal{R}} \sum_{n=1}^{N^{tr}} w_n \ell\left(\mathcal{R} \circ \Phi\left(G_n\right), \mathbf{Y}_n\right), \\
	\label{equation:objective2} \mathbf{W}^{*} &= {\rm argmin}_\mathbf{W} \sum_{1 \leq i < j \leq d}  \lVert \widehat{C}_{\mathbf{Z}_{*i}, \mathbf{Z}_{*j}}^{\mathbf{W}} \rVert_{\rm F}^2,
\end{align}
where $\ell$ denotes the cross-entropy loss for graph classification tasks or mean squared error loss for graph regression tasks. 
The optimization of graph weights $\mathbf{W}$ in Eq.~\eqref{equation:objective2} encourages the graph encoder to generate the graph representations $\mathbf{Z} = \Phi(\mathbf{G})$, where each dimension keeps independent with others and thus eliminates the spurious correlations. 
The optimization of graph encoder $\Phi$ and classifier $\mathcal{R}$ in Eq.~\eqref{equation:objective1} based on the weighted graph datasets will lead to good performance on the specific prediction tasks.

\subsection{Global-Local Graph Weight Estimator}

Note that directly optimizing Eqs.~\eqref{equation:objective1}\eqref{equation:objective2} requires all $N^{tr}$ graph weights $\mathbf{W}$ and graph representations $\mathbf{Z}$ to calculate accurately $\widehat{C}_{\mathbf{Z}_{*i}, \mathbf{Z}_{*j}}^{\mathbf{W}}$. Therefore, we need to load the entire dataset simultaneously for optimization, which is infeasible on large-scale datasets due to the high computational cost and excessive storage consumption.
A straightforward alternative is to learn only graph representations and corresponding weights over a mini-batch of data. However, the consistency of the weights cannot be maintained since different mini-batches do not share information. Therefore, the dependence between different graph representation dimensions is hard to eliminate over the whole training dataset.

To tackle this problem, we utilize a novel scalable global-local weight estimator to achieve the balance of optimization efficiency and weight consistency, inspired by~\cite{wu2018unsupervised, zhang2021deep, he2020momentum}.
In essence, we adopt global weights to keep the consistency of the learnable weights over the whole dataset, and the local weights encourage the independence of different dimensions of the graph representations over a mini-batch. Next, we elaborate on the detailed designs.

\textbf{Global weights.} 
We maintain $K$ groups of global representations $\mathbf{Z}^{(g)}=[\mathbf{Z}^{(g_1)}, \cdots, \mathbf{Z}^{(g_K)}]$ and the corresponding global weights $\mathbf{W}^{(g)}=[\mathbf{W}^{(g_1)}, \cdots, \mathbf{W}^{(g_K)}]$, where the size of each group equals to the mini-batch, i.e., $\mathbf{Z}^{(g_k)} \in \mathbb{R}^{|\mathcal{B}| \times d}$ and $\mathbf{W}^{(g_k)} \in \mathbb{R}^{| \mathcal{B}|}$.
They serve as the memory of the encoded graph representations and the corresponding weights from historical mini-batches during the training stage.
Since these global representations and weights are shared across different mini-matches, they maintain a global summarization of the whole training dataset. The size of global weights only depends on the mini-batch size, which is a hyper-parameter and independent of the training dataset size.  

\textbf{Local weights.}
For each mini-batch $\mathcal{B}$ of the input graphs $\{G_n\}_{n=1}^{|\mathcal{B}|}$, we first calculate their graph representations $\mathbf{Z}^{(l)}=\{\mathbf{Z}^{(l)}_{n*}\}_{n=1}^{|\mathcal{B}|}$, $\mathbf{Z}^{(l)}_{n*} = \Phi(G_n)$ and uniformly initialize the local graph weights, i.e., $\mathbf{W}^{(l)}=(1, 1, \dots, 1)$.
Then, the local graph representations $\mathbf{Z}^{(l)}$ and weights $\mathbf{W}^{(l)}$ are concatenated with the $K$ groups of global graph representations $\mathbf{Z}^{(g)}$ for optimization. 
We denote
\begin{equation}
\label{equ:concat}
\begin{aligned}
	\widehat{\mathbf{Z}} &=  \left[\mathbf{Z}^{(g_1)}, \cdots, \mathbf{Z}^{(g_K)} \mathbin\Vert \mathbf{Z}^{(l)}\right] \in \mathbb{R}^{(K + 1) |\mathcal{B}| \times d}, \\
	\widehat{\mathbf{W}} &= \left[\mathbf{W}^{(g_1)}, \cdots, \mathbf{W}^{(g_K)} \mathbin\Vert \mathbf{W}^{(l)}\right] \in \mathbb{R}^{(K + 1) |\mathcal{B}|},
\end{aligned}
\end{equation}
where $\left[\cdot \mathbin\Vert \cdot\right]$ is concatenation.
Then, we calculate the weighted partial cross-covariance matrix in Eq.~\eqref{equation:partial_cross_covariance_weight} using $\widehat{\mathbf{Z}}, \widehat{\mathbf{W}}$ and optimize the objective function. 
Using this estimator, the computational cost for each mini-batch is $O((K+1) |\mathcal{B}|)$, as opposed to  $O(N^{tr})$ in directly optimizing Eqs.~\eqref{equation:objective1}\eqref{equation:objective2}.

\textbf{Weights Update.}
At the end of each  training iteration,
we adopt a momentum update to dynamically update the global representations $\mathbf{Z}^{(g)}$ and weights $\mathbf{W}^{(g)}$ by the optimized local $\mathbf{Z}^{(l)}$ and $\mathbf{W}^{(l)}$:
\begin{equation}
\label{equation:update}
\begin{aligned}
	\mathbf{Z}^{(g_k)} &\leftarrow \gamma_k \mathbf{Z}^{(g_k)} + (1-\gamma_k) \mathbf{Z}^{(l)}, \\
	\mathbf{W}^{(g_k)} &\leftarrow \gamma_k \mathbf{W}^{(g_k)} + (1-\gamma_k) \mathbf{W}^{(l)}.
\end{aligned}
\end{equation}
Here $\gamma_k \in [0, 1)$ is a momentum coefficient for each group of global representations $\mathbf{Z}^{(g_k)}$ and $\mathbf{W}^{(g_k)}$ weights. 
The global $\mathbf{Z}^{(g_k)}$ and $\mathbf{W}^{(g_k)}$ with a large $\gamma_k$ serve as a long-term memory for global information over the whole training dataset, while those with a small $\gamma_k$ serve as a short-term memory.
Finally the global weights can be progressively updated and ensure the consistency of the whole graph dataset.

\subsection{Training Procedure}
The training procedure of our proposed \model is shown in Algorithm \ref{algo:myalgorithm}.

\begin{algorithm}[htp]
\caption{The training procedure of \mymodel.} 
\label{algo:myalgorithm}

\begin{flushleft}
\textbf{Input:} A graph dataset $\mathbf{G} = \{ G_n \}^{N}_{n=1}$ \\
\textbf{Output:} Learned graph encoder $\Phi^{*}$ and classifier $\mathcal{R}^{*}$ \\
\end{flushleft}
\begin{algorithmic}[1] 
\For{$e \gets 1$ to Epoch}
\For{sampled minibatch $\mathcal{B} = \{G_n\}_{n=1}^{|\mathcal{B}|}$}
    \State Calculate $\mathbf{Z}^{(l)}=\{\mathbf{Z}^{(l)}_{n*}\}_{n=1}^{|\mathcal{B}|}$, $\mathbf{Z}^{(l)}_{n*} = \Phi(G_n)$ 
    \State Initialize $\mathbf{W}^{(l)}=(1, 1, \dots, 1)$
    \State Concatenate global and local representations/weights as Eq.~\eqref{equ:concat}
    \For{$e^\prime \gets 1$ to Epoch\_Reweight}
		\State Optimize the graph weights by minimizing Eq.~\eqref{equation:objective2}
    \EndFor
	\State Back propagate with weighted prediction loss as Eq.~\eqref{equation:objective1}
    \State Update global representations and weights as Eq.~\eqref{equation:update}
\EndFor
\EndFor
\end{algorithmic}
\end{algorithm}

At the training stage, we iteratively optimize the graph weights $\mathbf{W}$, graph encoder $\Phi$, and classifier $\mathcal{R}$. Specifically, as shown in Algorithm \ref{algo:myalgorithm}, we first perform forward propagation for each sampled minibatch $\mathcal{B}$ to obtain the local graph representations $\mathbf{Z}^{(l)}=\{\mathbf{Z}^{(l)}_{n*}\}_{n=1}^{|\mathcal{B}|}$, $\mathbf{Z}^{(l)}_{n*} = \Phi(G_n)$ (line 3 in Algorithm \ref{algo:myalgorithm}) and uniformly initialize the local graph weights $\mathbf{W}^{(l)}=(1, 1, \dots, 1)$ (line 4).
To maintain consistency of the weights and improve efficiency, we concatenate the global representations and weights with local representations and weights to obtain $\widehat{\mathbf{Z}}$ and $\widehat{\mathbf{W}}$ (line 5).
After that, we calculate the partial cross-covariance matrix $\widehat{C}_{\widehat{\mathbf{Z}}_{*i}, \widehat{\mathbf{Z}}_{*j}}^{\widehat{\mathbf{W}}}$ and optimize the graph weights by minimizing the following objective function (line 7):
\begin{equation}
\begin{aligned}
	\mathbf{W}^{(l)*} &= {\rm argmin}_{\mathbf{W}^{(l)}} \sum_{1 \leq i < j \leq d}  \lVert \widehat{C}_{\widehat{\mathbf{Z}}_{*i}, \widehat{\mathbf{Z}}_{*j}}^{\widehat{\mathbf{W}}} \rVert_{\rm F}^2,
\end{aligned}
\end{equation}
Next, we optimize the graph encoder $\Phi$ and classifier $\mathcal{R}$ by performing back propagation with weighted prediction loss (line 9):
\begin{equation}
	\Phi^{*},  \mathcal{R}^{*} = {\rm argmin}_{\Phi,  \mathcal{R}} \sum_{n=1}^{|\mathcal{B}|} w_n \ell\left(\mathcal{R} \circ \Phi\left(G_n\right), \mathbf{Y}_n\right),
\end{equation}
where $w_n = {\mathbf{W}^{(l)*}_n}$ is the optimized weight for the $n$-$\rm{th}$ graph in the minibatch $\mathcal{B}$.
At the end of each iteration, the global representations and weights are updated by the optimized local graph representations and local graph weights (line 10).

At the testing stage, we directly adopt the optimized graph encoder $\Phi^{*}$ and classifier $\mathcal{R}^{*}$ to learn graph representations and conduct predictions.

\section{Experiments}
\label{section:experiments}
In this section, we empirically evaluate the effectiveness of the proposed \model on both synthetic and real-world datasets and conduct ablation studies. 
More experimental results (including hyper-parameter sensitivity, training dynamic, weight distribution, time complexity, etc.) are also present and analyzed in detail.

\subsection{Experimental Setup}

\subsubsection{Baselines}

We compare our \model with several representative state-of-the-art methods:
\begin{itemize}[leftmargin = 0.5cm]
	\item GCN~\cite{kipf2016semi}: It is one of the most famous GNNs, following a recursive neighborhood aggregation (or message passing) scheme.
	\item GIN~\cite{xu2018powerful}: It is shown to be one of the most expressive GNNs in representation learning of graphs.
	\item GCN-virtual and GIN-virtual~\cite{hu2020open}: We also consider the variants of GCN and GIN augmented with virtual node, i.e., adding a node that is connected to all the nodes in the original graphs.
	\item FactorGCN~\cite{yang2020factorizable}: It decomposes the input graph into several interpretable factor graphs for graph-level disentangled representations, which is a state-of-the-art disentangled GNN model for graph classification.
	\item PNA~\cite{corso2020principal}: It takes multiple neighborhood aggregation schemes into account and generalizes several GNN models with different neighborhood aggregation schemes.
    \item TopKPool~\cite{gao2019graph}: It propagates only part of the input and this part is not uniformly sampled from the input. It can thus select some local parts of the input graph and ignore the rest to summarize the graph representation.
    \item SAGPool~\cite{lee2019self}: It is a graph pooling method based on self-attention mechanism, which can be used to calculate attention scores and retain important nodes for graph-level representation.
\end{itemize}

\begin{table*}[t]
\small
\centering
\caption{The statistics of the datasets. \#Graphs is the number of graphs in the dataset. Average \#Nodes/\#Edges are the average number of nodes and edges in a graph of the dataset, respectively. \#Tasks denotes the dimensionality of output required for prediction. Task type includes binary classification, multi-classification, and regression. The various split methods for  training/validation/testing dataset cover complex and realistic distribution shifts.}
\label{table:statistic}
\begin{adjustbox}{max width=\textwidth}
\begin{tabular}{|l|l|r|r|r|r|c|c|c|}	
\hline
\textbf{Category}                                                                         & \textbf{Name} & \textbf{\#Graphs} & \textbf{\begin{tabular}[c]{@{}c@{}}Average \\ \#Nodes\end{tabular}} & \textbf{\begin{tabular}[c]{@{}c@{}}Average\\ \#Edges\end{tabular}} & \textbf{\#Tasks} & \textbf{\begin{tabular}[c]{@{}c@{}}Task\\ Type\end{tabular}} & \textbf{\begin{tabular}[c]{@{}c@{}}Split\\ Method\end{tabular}} & \textbf{Metric} \\
\hline
\multirow{2}{*}{Synthetic}                      & TRIANGLES  & 4,000   & 15.6              & 48.9              & 1      & Regression    & Size         & Accuracy \\
                                                & MNIST-75SP & 7,000   & 66.8              & 600.2             & 1      & Multi-class.  & Feature      & Accuracy \\
\hline
\multirow{3}{*}{\begin{tabular}[c]{@{}l@{}}Molecule   and \\ social datasets\end{tabular}} & COLLAB     & 5,000   & 74.5              & 2457.8            & 1      & Multi-class.  & Size         & Accuracy \\
                                                & PROTEINS   & 1,113   & 39.1              & 72.8              & 1      & Binary class. & Size         & Accuracy \\
                                                & D\&D        & 1,178   & 284.3             & 715.7             & 1      & Binary class. & Size         & Accuracy \\
\hline
\multirow{12}{*}{\begin{tabular}[c]{@{}l@{}}Open Graph Benchmark\\ OGBG-MOL*\end{tabular}}          & TOX21      & 7,831   & 18.6              & 19.3              & 12     & Binary class. & Scaffold     & ROC-AUC  \\
                                                & BACE       & 1,513   & 34.1              & 36.9              & 1      & Binary class. & Scaffold     & ROC-AUC  \\
                                                & BBBP       & 2,039   & 24.1              & 26.0              & 1      & Binary class. & Scaffold     & ROC-AUC  \\
                                                & CLINTOX    & 1,477   & 26.2              & 27.9              & 2      & Binary class. & Scaffold     & ROC-AUC  \\
                                                & SIDER      & 1,427   & 33.6              & 35.4              & 27     & Binary class. & Scaffold     & ROC-AUC  \\
                                                & TOXCAST    & 8,576   & 18.8              & 19.3              & 12     & Binary class. & Scaffold     & ROC-AUC  \\
                                                & HIV        & 41,127  & 25.5              & 27.5              & 1      & Binary class. & Scaffold     & ROC-AUC  \\
                                                & ESOL       & 1,128   & 13.3              & 13.7              & 1      & Regression    & Scaffold     & RMSE     \\
                                                & FREESOLV   & 642     & 8.7               & 8.4               & 1      & Regression    & Scaffold     & RMSE     \\
\hline
\end{tabular}
\end{adjustbox}
\end{table*}

\subsubsection{Datasets}
\label{section:datasets}
To cover more realistic and challenging cases of graph distribution shifts, we compare our method and baselines on both synthetic and real-world datasets:
\begin{itemize}[leftmargin = 0.5cm]
    \item \textbf{Synthetic Datasets.} 
    
    We use two synthetic datasets to evaluate the effectiveness of our proposed method, and examples of these datasets are shown in Figure~\ref{fig:example_triangles} and~\ref{fig:example_mnist}.
    
    (1) \textbf{TRIANGLES}. Counting the number of triangles in a graph is a common task that can be solved analytically but is challenging for GNNs. We first generate 4,000 random graphs, and train on graphs containing $4$ to $25$ nodes, and test on graphs with $4$ to $100$ nodes. The node features are set as one-hot degrees. The dataset is split into 3,000/500/500 graphs used as training/validation/testing sets. The task is to predict the number of triangles in each graph. The number of classes is 10 (i.e., each graph has 1 to 10 triangles). Based on this setting, there exist distribution shifts with regard to graph sizes between training and testing data. 
    
    (2) \textbf{MNIST-75SP}. Each graph in MNIST-75SP is converted from an image in MNIST~\cite{lecun1998gradient} using super-pixels~\cite{achanta2012slic}. We randomly sample 7,000 images of MNIST and extract no more than $75$ superpixels for each image to generate the graph. The node features are set as the super-pixel coordinates and intensity. The dataset is split into 6,000/500/500 graphs used as training/validation/testing sets. The task is to classify each graph into the corresponding handwritten digit labeled from $0$ to $9$. To simulate distribution shifts with respect to graph features, we follow~\cite{knyazev2019understanding} and generate two testing graph datasets. For the first testing set, Test(noise), we add Gaussian noise, drawn from $\mathcal{N} (0, 0.4)$, to node features. For the second testing set, Test(color), we colorize images by adding two more channels and add independent Gaussian noise, drawn from $\mathcal{N} (0, 0.4)$, to each channel. The graph structures (adjacency matrices) are not changed for testing graphs.

    \item \textbf{Real-world Datasets.} 
    
    (1) \textbf{Molecule and social datasets}. We consider three commonly used graph classification benchmarks: \textbf{COLLAB}~\cite{shrivastava2014new}, \textbf{PROTEINS}~\cite{borgwardt2005protein}, and \textbf{D\&D}~\cite{dobson2003distinguishing}.  Following \cite{knyazev2019understanding}, these datasets are split based on the size of each graph. 
    \textbf{D\&D$\bm{_{200}}$} and \textbf{D\&D$\bm{_{300}}$} denote the two datasets whose maximum graph size in the training set is 200 and 300, respectively.
    All the methods are trained on smaller graphs and tested on unseen larger graphs. Specifically, \textbf{COLLAB} is derived from 3 public collaboration datasets, i.e., High Energy Physics, Condensed Matter Physics, and Astro Physics. We train on graphs with $32$ to $35$ nodes and test on graphs with $32$ to $492$ nodes. \textbf{PROTEINS} is a protein dataset. We train on graphs with $4$ to $25$ nodes and test on graphs with $6$ to $620$ nodes. \textbf{D\&D} is also a dataset that consists of proteins. We consider two types of splitting methods, termed \textbf{D\&D$\bm{_{200}}$} and \textbf{D\&D$\bm{_{300}}$}. For \textbf{D\&D$\bm{_{200}}$}, we train on graphs with $30$ to $200$ nodes and test on graphs with $201$ to $5,748$ nodes. For \textbf{D\&D$\bm{_{300}}$}, we train on 500 graphs with $30$ to $300$ nodes and test on other graphs with $30$ to $5,748$ nodes.
    
    (2) \textbf{Open Graph Benchmark (OGB)}~\cite{hu2020open}. We consider 9 graph property prediction datasets from a benchmark of distribution shifts \textbf{OGBG-MOL$*$} in Open Graph Benchmark (OGB), i.e., TOX21, BACE, BBBP, CLINTOX, SIDER, TOXCAST, HIV, ESOL, FREESOLV. The task is to predict the target molecular properties as accurately as possible. We adopt the default scaffold splitting procedure, namely splitting the graphs based on their two-dimensional structural frameworks. Note that this scaffold splitting strategy aims to separate structurally different molecules into different subsets, which provides a more realistic and challenging scenario of out-of-distribution generalization. Figure~\ref{fig:example_ogb} shows some examples of the dataset.

\end{itemize}

For the synthetic datasets in the experiments:
\begin{itemize}[leftmargin = 0.5cm]
\item TRIANGLES: Each graph in this dataset is a random graph.
\item MNIST-75SP: It is generated from MNIST~\cite{lecun1998gradient}: \url{http://yann.lecun.com/exdb/mnist/} with license unspecified.
\end{itemize}

The real-world datasets are publicly available.
\begin{itemize}[leftmargin = 0.5cm]
\item COLLAB, PROTEINS, D\&D: \url{https://chrsmrrs.github.io/datasets/} with license unspecified
\item Open Graph Benchmark (OGB): \url{https://ogb.stanford.edu/docs/graphprop/} with MIT License
\end{itemize}

\subsubsection{Implementation Details}
We implement our method in PyTorch. 
The number of epochs (i.e., Epoch in Algorithm \ref{algo:myalgorithm}) is set to 100.
The batch size is chosen from \{64, 128, 256\}.
The learning rate is chosen from \{0.0001, 0.001\}.
The number of epochs of learning graph weights (i.e., Epoch\_Reweight in Algorithm \ref{algo:myalgorithm}) is set to 20. 
The dimensionality of the representations and hidden layers $d$ is chosen from \{128, 300\} for Open Graph Benchmark, and \{64, 256\} for other datasets.  
We use GIN~\cite{xu2018powerful} as the graph encoder $\Phi: \mathcal{G} \rightarrow \mathcal{Z}$  since it is shown to be one of the most expressive GNNs, and the number of layers is chosen from [2, 6].
We set $Q=1$ to sample random Fourier features.
The $\ell^2$-norm is adopted on the weights to prevent degenerated solutions.
The number of groups of global representations and weights $K=1$ with the momentum coefficient $\gamma = 0.9$ in the updating step.
The classifier $\mathcal{R}: \mathcal{Z} \rightarrow \mathcal{Y}$ is realized by a two-layer MLP.
These hyper-parameters are tuned on the validation set. 
We report the mean values with standard deviations of 10 repeated experiments.

We conduct the experiments with the following hardware and software configurations:
\begin{itemize}
\item Operating System: Ubuntu 18.04.1 LTS
\item CPU: Intel(R) Xeon(R) CPU E5-2699 v4@2.20GHz
\item GPU: NVIDIA GeForce GTX TITAN X with 12GB of Memory
\item Software: Python 3.6.5; NumPy 1.18.0; PyTorch 1.7.0; PyTorch Geometric 1.6.1.
\end{itemize}

\begin{table*}[t]
\small
\centering
\caption{Graph classification accuracy (\%) on training and testing sets of two synthetic datasets. 
Test(large) denotes larger graph sizes in testing set and Test(noise)/Test(color) represent adding Gaussian noises/color noises respectively. In each column, the boldfaced score denotes the best result and the underlined score represents the second-best result. $\pm$ denotes standard deviation.}
\label{table:synthetic}
\begin{adjustbox}{max width=0.99\textwidth}
\begin{tabular}{|c|cc|ccc|}
\hline
                  & \multicolumn{2}{c|}{\textbf{TRIANGLES}}        & \multicolumn{3}{c|}{\textbf{MNIST-75SP}}                                \\
\hline
            & Train                 & Test(large)                  & Train                  & Test(noise)           & Test(color)           \\
\hline
GCN               & 28.3$\pm$0.6          & 21.3$\pm$1.9          & 51.7$\pm$1.0           & 26.5$\pm$1.4          & 27.0$\pm$1.3          \\
GCN-virtual       & 32.4$\pm$0.6          & 17.0$\pm$1.8          & 55.1$\pm$2.3           & 26.0$\pm$1.5          & 26.1$\pm$1.8          \\
GIN               & {\ul34.7$\pm$0.7}          & 22.2$\pm$1.9          & {\ul 67.6$\pm$0.8}           & {\ul 27.9$\pm$2.5}    & {\ul 34.3$\pm$4.4}    \\
GIN-virtual       & 34.2$\pm$0.6          & 17.6$\pm$1.7          & 66.7$\pm$0.9           & 25.7$\pm$2.9          & 33.4$\pm$1.2          \\
FactorGCN         & 10.6$\pm$1.6          & 4.2$\pm$0.9           & 46.7$\pm$1.2           & 19.7$\pm$1.4          & 24.8$\pm$1.3          \\
PNA               & \textbf{43.7$\pm$3.6} & 16.8$\pm$2.4          & \textbf{83.0$\pm$0.9}  & 22.8$\pm$7.3          & 29.2$\pm$6.3          \\
TopKPool          & 28.3$\pm$0.3          & 22.0$\pm$0.2          & 61.0$\pm$3.7           & 17.0$\pm$1.0          & 16.9$\pm$1.5          \\
SAGPool           & 26.7$\pm$1.0          & {\ul 23.7$\pm$0.7}    & 60.2$\pm$1.3           & 19.6$\pm$3.4          & 20.1$\pm$3.7          \\
\textbf{\mymodel} & 29.9$\pm$0.7    & \textbf{25.1$\pm$0.8} & 63.2$\pm$1.1     & \textbf{31.5$\pm$0.9} & \textbf{38.5$\pm$1.5} \\

\hline
\end{tabular}
\end{adjustbox}
\end{table*}

\subsection{Results on Synthetic Graphs}

The results on TRIANGLES and MNIST-75SP are reported in Table~\ref{table:synthetic}.
On TRIANGLES, there exist distribution shifts on the graph sizes.
\model consistently achieves the best testing performance compared with other baselines on the out-of-distribution testing graphs, demonstrating the OOD generalization capability of our method.
The accuracy of a strong baseline PNA on training graphs is impressive but drops significantly on the OOD testing graphs.
FactorGCN, as a disentangled graph representation learning method, decomposes the input graph into several independent factor graphs so that it may change the semantic implication of representations into these implicit factors and affect the performance.
In contrast, \model learns graph weights so that the semantic of the graph representations will not be affected, leading to better generalization ability.

On MNIST-75SP, there exist distribution shifts on the graph features, i.e., graphs in the testing datasets have larger noises.
\model achieves the best performance consistently compared with other methods.
For this dataset, each graph consists of super-pixel nodes and edges that are
formed based on the spatial distance between super-pixel centers. Therefore, the graph topological structures are relatively more discriminative than node features in making predictions.
Traditional GNNs fuse heterogeneous information from both graph topological structures and features into unified graph representations, so these baselines may learn the spurious correlations, leading to poor generalization performance.
As the complex non-linear dependencies between graph structures and features are eliminated, our method is able to learn the true connections between relevant representations (i.e., informative graph topological structures) and labels, and conduct inference according to them only, thus generalize better.

\begin{table*}[t]
\small
\centering
\caption{Graph classification accuracy (\%) on the \textbf{testing set} of \model and baselines. 
Our \model outperforms the baselines significantly on all graph classification benchmarks, indicating its superiority against graph size distribution shifts. The best result and the second-best result for each dataset are in bold and underlined, respectively.
}
\label{table:performance_realworld}
\begin{adjustbox}{max width=0.999\textwidth}
\begin{tabular}{|c|c|c|c|c|}
\hline
                    & \textbf{COLLAB$\bm{_{35}}$}       & \textbf{PROTEINS$\bm{_{25}}$}     & \textbf{D\&D$\bm{_{200}}$}      & \textbf{D\&D$\bm{_{300}}$}      \\
\hline
\# Train/Test graphs & 500/4500              & 500/613               & 462/716               & 500/678               \\
\#Nodes Train        & 32-35                 & 4-25                  & 30-200                & 30-300                \\
\#Nodes Test         & 32-492                & 6-620                 & 201-5748              & 30-5748               \\
\hline
GCN                 & 65.9$\pm$3.4          & 75.1$\pm$2.2          & 29.2$\pm$8.2          & 71.9$\pm$3.6          \\
GCN-virtual         & 61.5$\pm$1.6          & 70.4$\pm$3.7          & 41.6$\pm$8.0          & 71.6$\pm$4.4          \\
GIN                 & 55.5$\pm$4.9          & 74.0$\pm$2.7          & 43.0$\pm$8.3          & 67.8$\pm$4.3          \\
GIN-virtual         & 54.8$\pm$2.7          & 66.0$\pm$7.5          & 46.7$\pm$4.5         & 72.1$\pm$4.3          \\
FactorGCN           & 51.0$\pm$1.3          & 63.5$\pm$4.8          & 42.3$\pm$3.1          & 55.9$\pm$1.6          \\
PNA                 & 59.6$\pm$5.5          & 71.4$\pm$3.4          & 47.3$\pm$6.8         & 70.1$\pm$2.1          \\
TopKPool            & 52.8$\pm$1.0          & 64.9$\pm$3.0          & 34.6$\pm$5.6          & 69.3$\pm$3.6          \\
SAGPool             & {\ul 67.0$\pm$1.7}    & {\ul 76.2$\pm$0.7}    & {\ul 54.3$\pm$5.0}    & {\ul 78.4$\pm$1.1}     \\
\mymodel            & \textbf{67.2$\pm$1.8} & \textbf{78.4$\pm$0.9} & \textbf{60.3$\pm$4.5} & \textbf{80.1$\pm$1.0} \\
\hline
\end{tabular}
\end{adjustbox}
\end{table*}

\begin{table*}[t]
\small
\centering
\caption{Results on nine Open Graph Benchmark (OGB) datasets. We report the ROC-AUC (\%) for classification tasks and RMSE for regression tasks with the standard deviation on the \textbf{test set} of all methods. None of the baseline methods is consistently competitive across all datasets, while our proposed method shows impressive performance. ($\uparrow$) means that higher values indicate better results, and ($\downarrow$) represents the opposite.}
\label{table:performance_ogb}
\begin{adjustbox}{max width=0.99\textwidth}
\begin{tabular}{|c|c|c|c|c|c|c|c||c|c|}
\hline
     & \textbf{TOX21}        & \textbf{BACE}         & \textbf{BBBP}         & \textbf{CLINTOX}      & \textbf{SIDER}        & \textbf{TOXCAST}      & \textbf{HIV}  & \textbf{ESOL}          & \textbf{FREESOLV}              \\
\hline
Metric & \multicolumn{7}{c||}{{ROC-AUC ($\uparrow$)}} & \multicolumn{2}{c|}{{RMSE ($\downarrow$)}}\\
\hline
GCN         & 75.3$\pm$0.7          & {\ul 79.2$\pm$1.4}    & 68.9$\pm$1.5          & {\ul 91.3$\pm$1.7}   & 59.6$\pm$1.8          & 63.5$\pm$0.4          & 76.1$\pm$1.0   & 1.11$\pm$0.03          & 2.64$\pm$0.24       \\
GCN-virtual & 77.5$\pm$0.9          & 68.9$\pm$7.0          & 67.8$\pm$2.4          & 88.6$\pm$2.1          & 59.8$\pm$1.5          & {\ul 66.7$\pm$0.5}    & 76.0$\pm$1.2   & 1.02$\pm$0.10          & 2.19$\pm$0.12       \\
GIN         & 74.9$\pm$0.5          & 73.0$\pm$4.0          & 68.2$\pm$1.5          & 88.1$\pm$2.5          & 57.6$\pm$1.4          & 63.4$\pm$0.7          & 75.6$\pm$1.4   & 1.17$\pm$0.06          & 2.76$\pm$0.35       \\
GIN-virtual & {\ul 77.6$\pm$0.6}    & 73.5$\pm$5.2          & {\ul 69.7$\pm$1.9}    & 84.1$\pm$3.8          & 57.6$\pm$1.6          & 66.1$\pm$0.5          & 77.1$\pm$1.5   & 1.00$\pm$0.07          & 2.15$\pm$0.30       \\
FactorGCN   & 57.8$\pm$2.1          & 70.0$\pm$0.6          & 54.1$\pm$1.1          & 64.2$\pm$2.1          & 53.3$\pm$1.7          & 51.2$\pm$0.8          & 57.1$\pm$1.5   & 3.39$\pm$0.15          & 5.69$\pm$0.32      \\
PNA         & 71.5$\pm$0.5          & 77.4$\pm$2.1          & 66.2$\pm$1.2          & 81.2$\pm$2.0          & 59.6$\pm$1.1          & 60.6$\pm$0.2          & {\ul 79.1$\pm$1.3}  & {\ul 0.94$\pm$0.02}    & 2.92$\pm$0.16   \\
TopKPool    & 75.6$\pm$0.9          & 76.9$\pm$2.4          & 68.6$\pm$1.1          & 86.9$\pm$1.1          & 60.6$\pm$1.5          & 64.7$\pm$0.1          & 76.7$\pm$1.1    & 1.17$\pm$0.03          & {\ul 2.08$\pm$0.10}      \\
SAGPool     & 74.7$\pm$3.1          & 76.6$\pm$1.0          & 69.3$\pm$2.1          & 88.7$\pm$1.0          & {\ul 61.3$\pm$1.3}    & 64.8$\pm$0.2          & 77.7$\pm$1.3    & 1.22$\pm$0.05          & 2.28$\pm$0.12      \\
\mymodel    & \textbf{78.4$\pm$0.8} & \textbf{81.3$\pm$1.2} & \textbf{70.1$\pm$1.0} & \textbf{91.4$\pm$1.3} & \textbf{64.0$\pm$1.3} & \textbf{68.7$\pm$0.3} & \textbf{79.5$\pm$0.9} & \textbf{0.88$\pm$0.05} & \textbf{1.81$\pm$0.14} \\
\hline
\end{tabular}
\end{adjustbox}
\end{table*}

\begin{figure*}[t]
\centering
\begin{subfigure}{.3\linewidth}

  \centering
  \includegraphics[width=.99\linewidth]{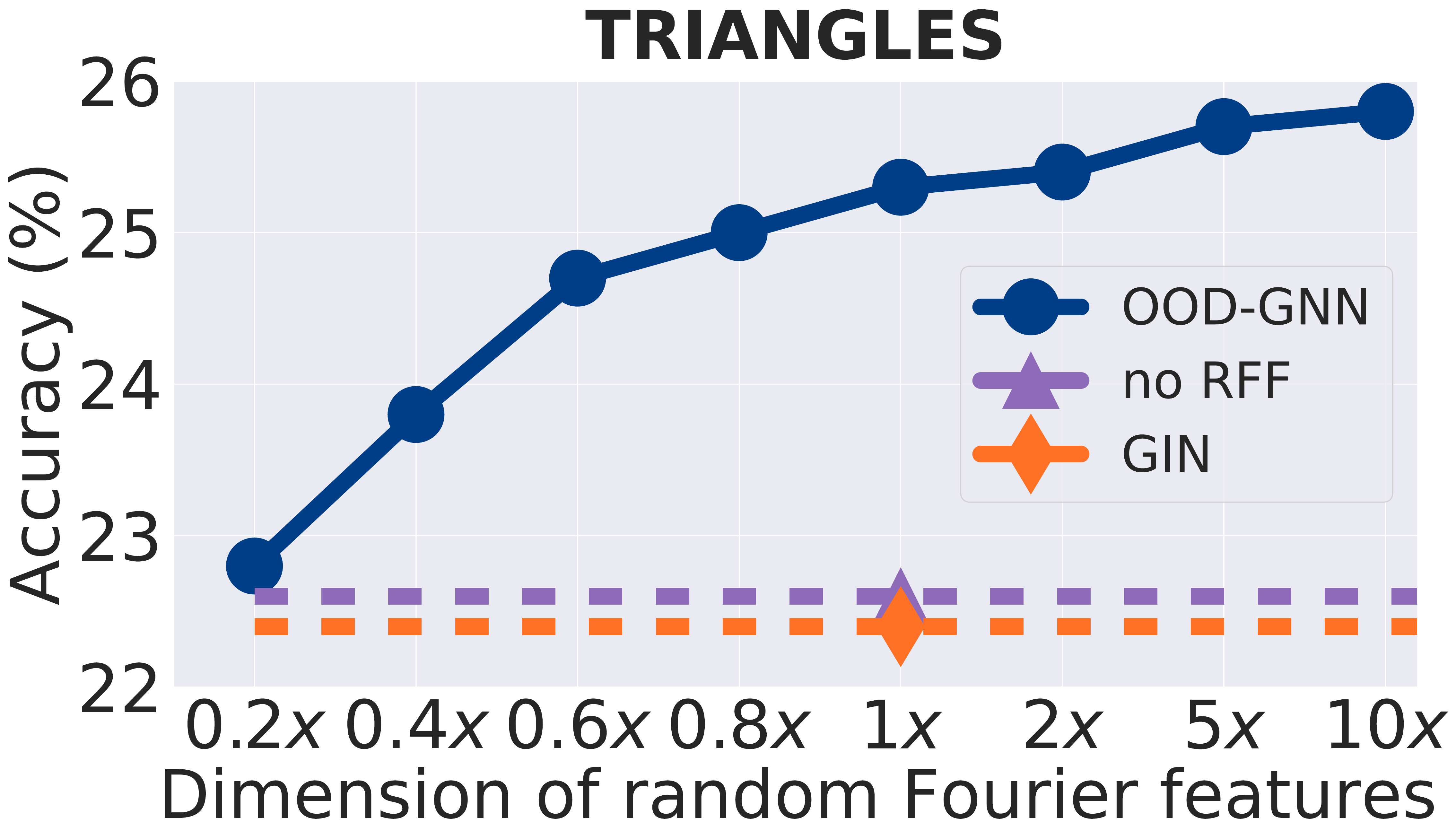} 
  \caption{TRIANGLES}
\end{subfigure}
\begin{subfigure}{.3\linewidth}

  \centering
  \includegraphics[width=.99\linewidth]{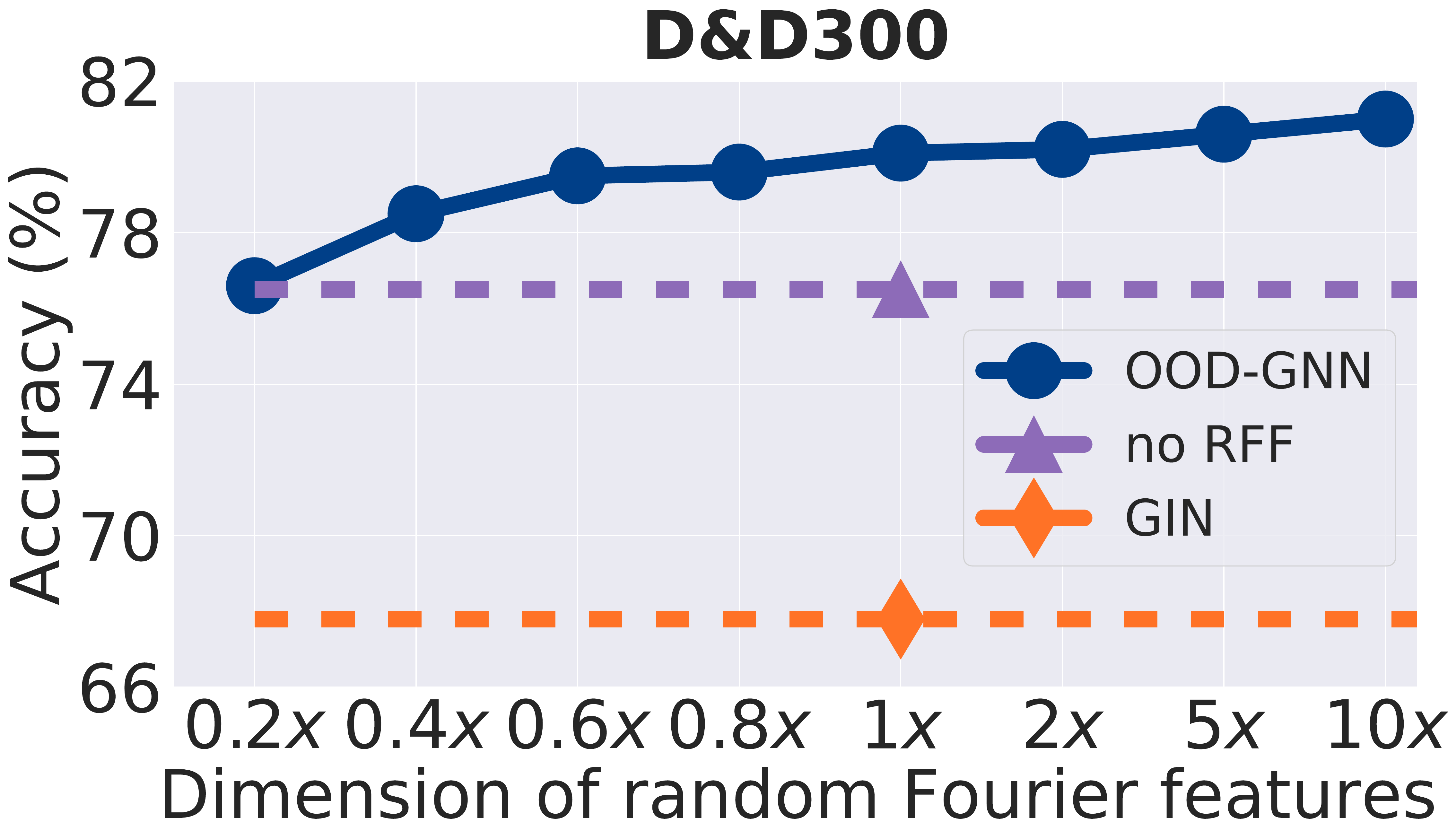} 
  \caption{D\&D$_{300}$}
\end{subfigure}
\begin{subfigure}{.3\linewidth}
  \centering
  \includegraphics[width=.99\linewidth]{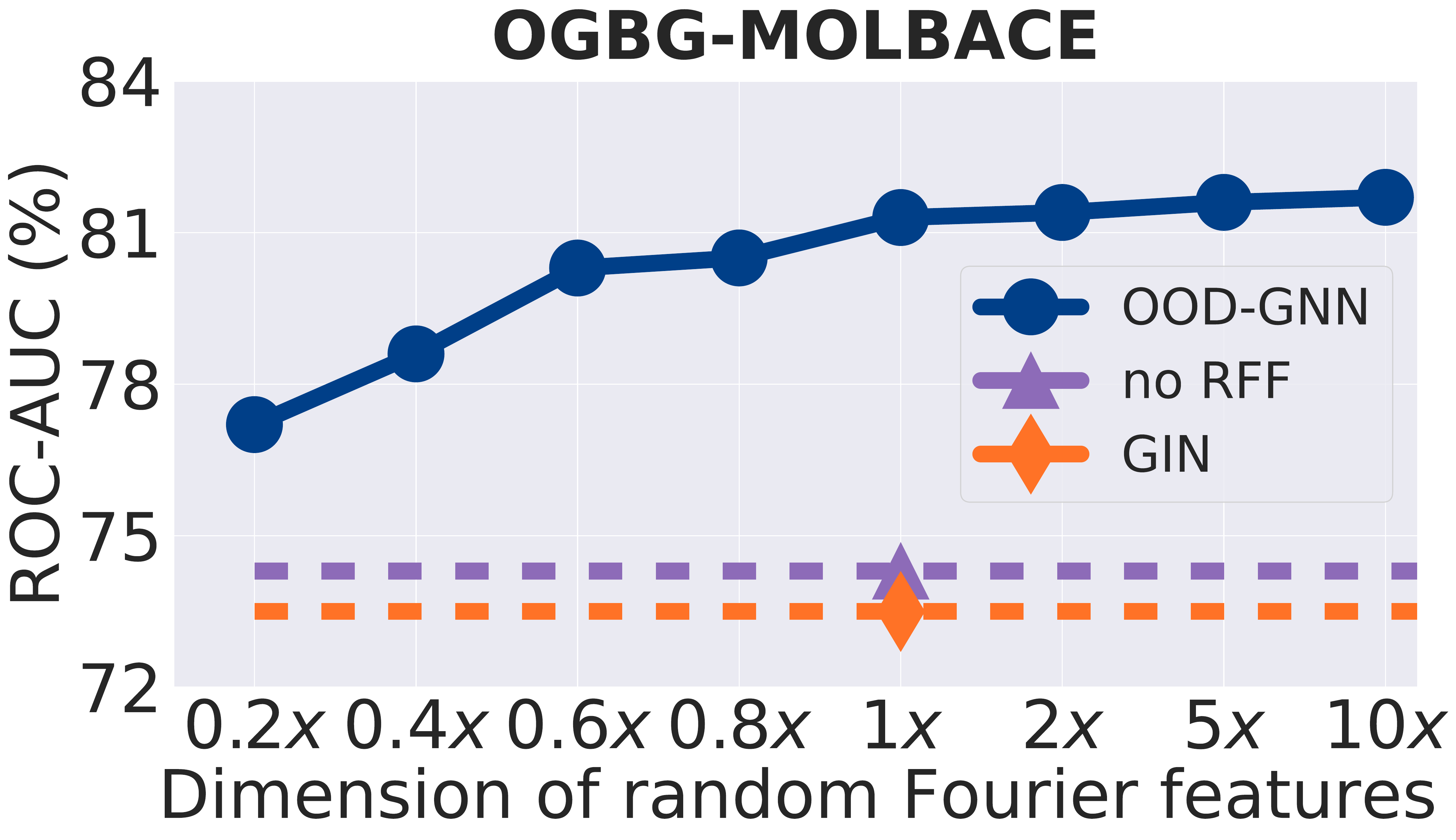} 
  \caption{OGBG-MOLBACE}
\end{subfigure}
\caption{
Ablation study results of our method. The blue curves with circle markers show that  as dimensionality of random Fourier features increases, the generalization performance of \model improves. The purple markers show that if we remove random Fourier features and only eliminate linear correlation, the performance drops significantly. The orange markers represent the results of GIN, the graph encoder baseline in our method. 
}
\label{figure:ablation}
\end{figure*}

\subsection{Results on Real-world Graphs}
On real-world molecule and social datasets (i.e., COLLAB, PROTEINS, and D\&D), the training and testing graphs are split by graph sizes, i.e., our method and baselines are trained on small graphs and tested on larger graphs.
The results are presented in Table~\ref{table:performance_realworld}. 
\model consistently yields the best testing performance on all the datasets. In particular, \model improves over the strongest baselines by 2.2\%, 6.0\%, and 1.7\% in PROTEINS${_{25}}$, D\&D${_{200}}$, D\&D${_{300}}$ respectively.
Our model achieves the best out-of-distribution generalization performance under size distribution shifts by encouraging independence between relevant and irrelevant representations.
The results of baselines degrade due to the spurious correlations between irrelevant representations and labels. 
For example, each graph in the COLLAB dataset corresponds to an ego-network of different researchers from one field, and the label denotes the corresponding research field. The truly predictive representations are from the graph topological structures. If the GNN models learn spurious correlations between graph sizes and labels, they will fail to make correct predictions on larger OOD testing graphs.

The graph classification results on nine Open Graph Benchmark (OGB) datasets are shown in Table~\ref{table:performance_ogb}.
The datasets are split based on the scaffold, i.e., the two-dimensional structural framework.
So the distribution shifts between training and testing graphs exist on the graph topological structure and features, leading to a more challenging scenario.
None of the baselines is consistently competitive across all datasets, as opposed to our proposed method.
Notice that adding virtual nodes to GCN and GIN is not a promising improvement for generalization since it can provide performance gains on some datasets but fail on the others. 
FactorGCN shows poor results on these datasets, possibly because it enforces the decomposition of the input graphs into several independent factor graphs for disentanglement, which is hard to achieve without sufficient supervision.
PNA is proposed to address the size generalization problem but still fails under the more complex distribution shifts.
TopKPool selects some local parts of the input graph and ignores the others. The strongest baseline on molecule and social datasets, i.e., SAGPool, pools the nodes with the self-attention mechanism. However, the accurate selection for TopKPool and calculation of attention scores for SAGPool are easily affected by the spurious correlations on OOD test graphs and therefore also fail to generalize.
In contrast, \model shows a strong capability of out-of-distribution generalization when the input graphs have complicated structures, especially for the large-scale real-world graphs.

\subsection{Ablation Studies}

We perform ablation studies over a number of key components of our method to analyze their functionalities more deeply. 
Specifically, we compare \model with the following two variants: 
(1) \textbf{Variant 1}: it sets the dimensionality of random Fourier features to different values. (2) \textbf{Variant 2}: it removes all the random Fourier features. For simplicity, we only report the results on one synthetic dataset (i.e., TRIANGLES) and two real-world datasets (i.e., D\&D${_{300}}$ and OGBG-MOLBACE), while the results on other datasets show similar patterns.

\textbf{Variant 1} exploits the effect of different dimensions of random Fourier features. 
Note that our method adopts random Fourier features (see Eq.~\eqref{equation:rff}), which sample from Gaussian to learn the graph weights and encourage the independence of representations.
It is shown in~\cite{strobl2017approximate} that if sampling more random Fourier features (i.e., when $Q$ in Eq.~\eqref{equation:rff} increases), the learned graph representations will be more independent. However, there exists a trade-off between independence and computational efficiency since the more random Fourier features are sampled, the higher the computational cost becomes. 
When the computational resources are extremely limited, it is also feasible to randomly select part of the dimensions in graph representations to calculate the dependence.
In Figure~\ref{figure:ablation}, the x-axis represents the dimensionality of random Fourier features compared to graph representations, e.g., "2x" indicates $Q=2$ in Eq.~\eqref{equation:rff}, while "0.2x" means we randomly select 20\% dimensions of graph representations.
We observe from Figure~\ref{figure:ablation} that as the dimensionality of random Fourier features increases, the performance on OOD testing graphs grows consistently, which 
demonstrates that eliminating the statistical dependence between different dimensions of the graph representations will encourage the independence between relevant and irrelevant representations and lead to better out-of-distribution generalization ability.

\textbf{Variant 2} removes all the random Fourier features and the optimization in Eqs.~\eqref{equation:objective1}\eqref{equation:objective2} will degenerate to linear cases, i.e., only eliminating linear correlation rather than encouraging independence between different dimensions of graph representations.
In Figure~\ref{figure:ablation}, this variant is termed as "no RFF". 
We can observe a clear performance drop for this variant, demonstrating that the complex non-linear dependencies are common in the graph representations.
By eliminating non-linear dependence between representations, the GNNs will be encouraged to learn true connections between the input graphs and the corresponding labels.

\subsection{Training Dynamic}
\begin{figure*}[t]
\centering
    \begin{subfigure}[t]{0.32\textwidth}
        \centering
        \includegraphics[width=.99\textwidth]{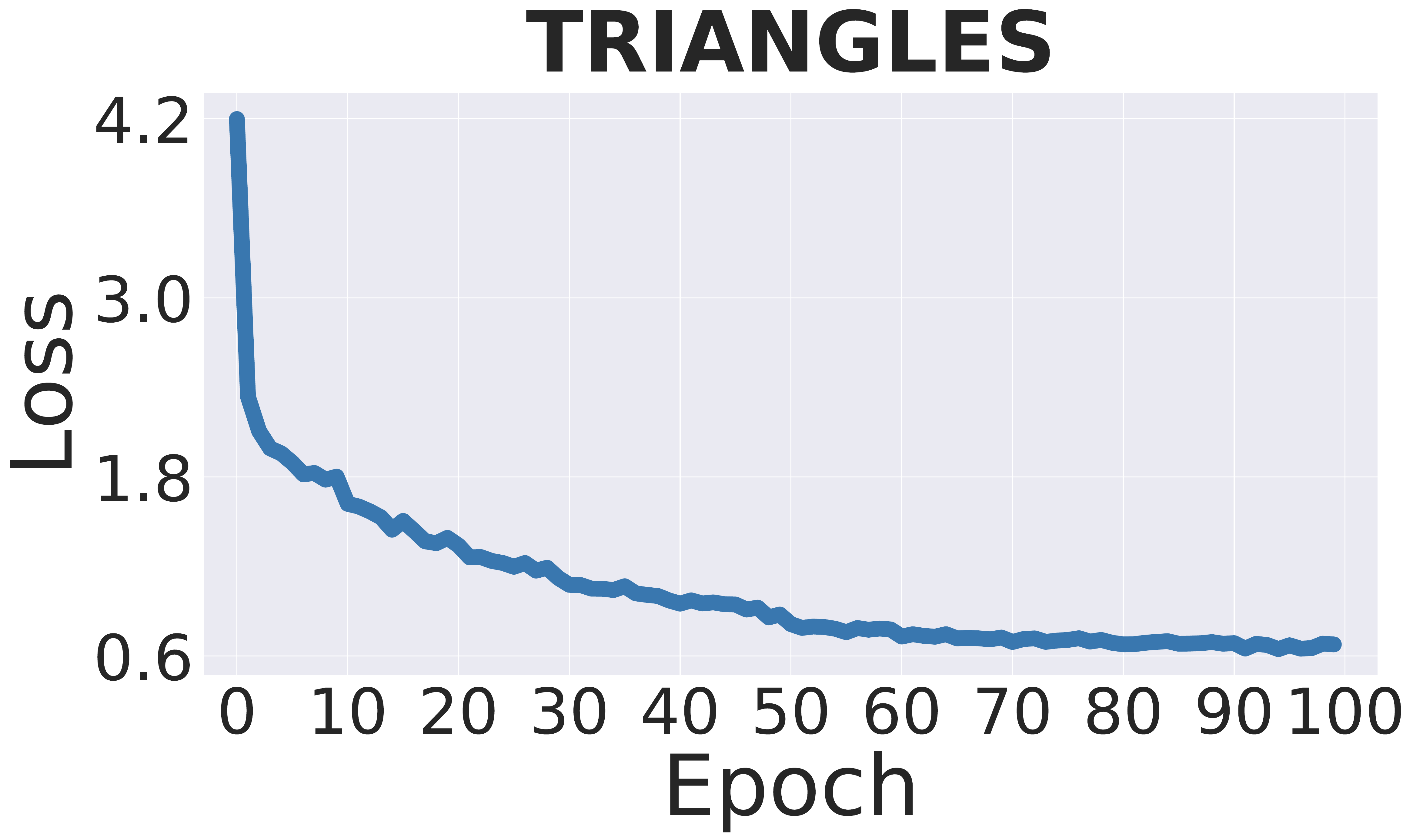}
        \caption{Training loss on TRIANGLES.}
    \end{subfigure}
    ~ 
    \begin{subfigure}[t]{0.32\textwidth}
        \centering
        \includegraphics[width=.99\textwidth]{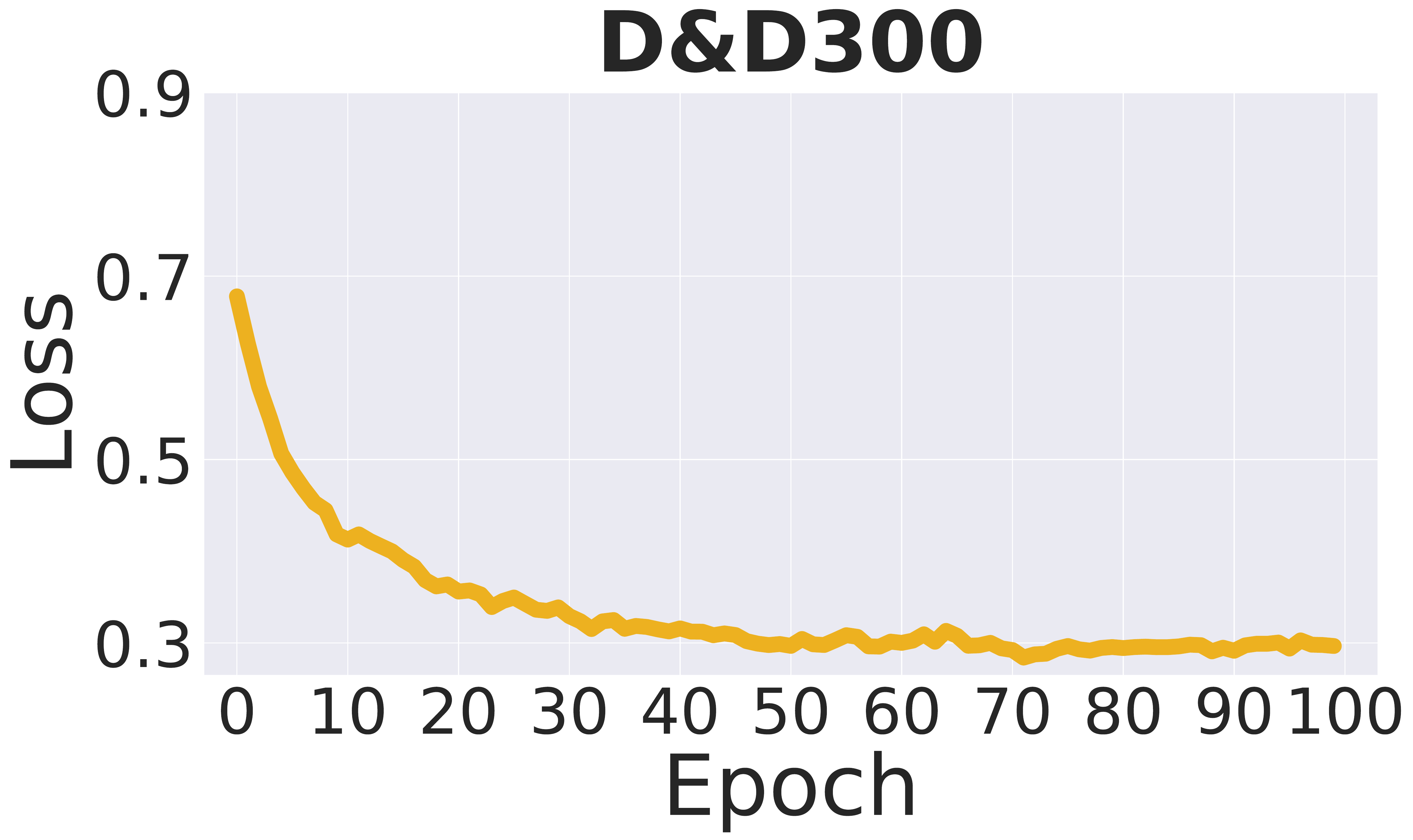}
        \caption{Training loss on D\&D$_{300}$.}
    \end{subfigure}
    ~ 
    \begin{subfigure}[t]{0.32\textwidth}
        \centering
        \includegraphics[width=.99\textwidth]{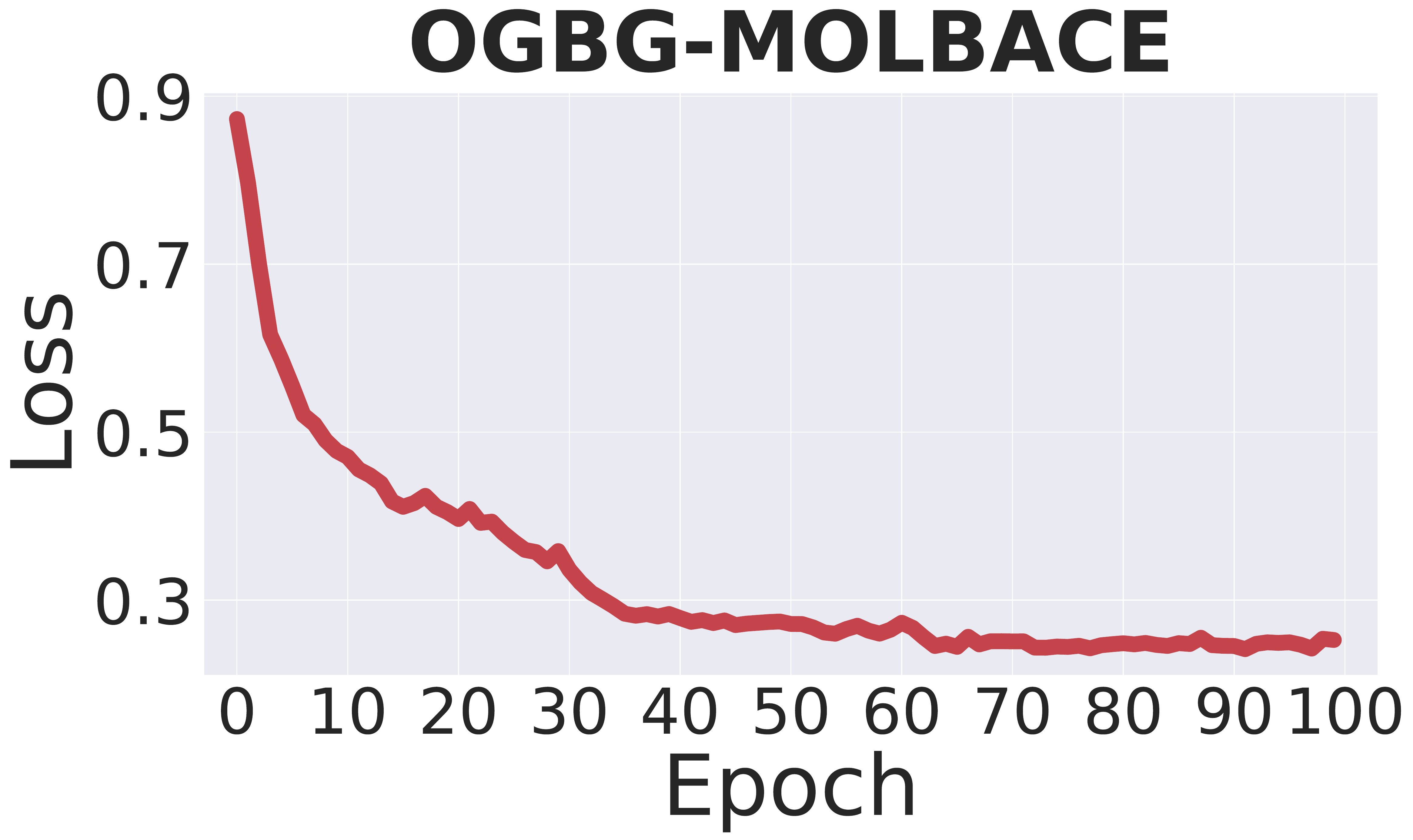}
        \caption{Training loss on OGBG-MOLBACE.}
    \end{subfigure}
\caption{The weighted prediction loss in the training process on three datasets.}
\label{figure:training_loss}
\end{figure*}

We can observe the convergence of our proposed method empirically, although Eqs.~\eqref{equation:objective1}\eqref{equation:objective2} are iteratively optimized.
In Figure~\ref{figure:training_loss} (a)(b)(c), we show the weighted prediction loss in the training process on TRIANGLES, D\&D$_{300}$, and OGBG-MOLBACE, respectively. 
The loss converges in no more than 100 epochs to about 0.67, 0.30, and 0.25 on the three datasets, respectively. 
The results on the other datasets show similar patterns.

\begin{figure*}[t]
\centering
    \begin{subfigure}[t]{0.32\textwidth}
        \centering
        \includegraphics[width=.99\textwidth]{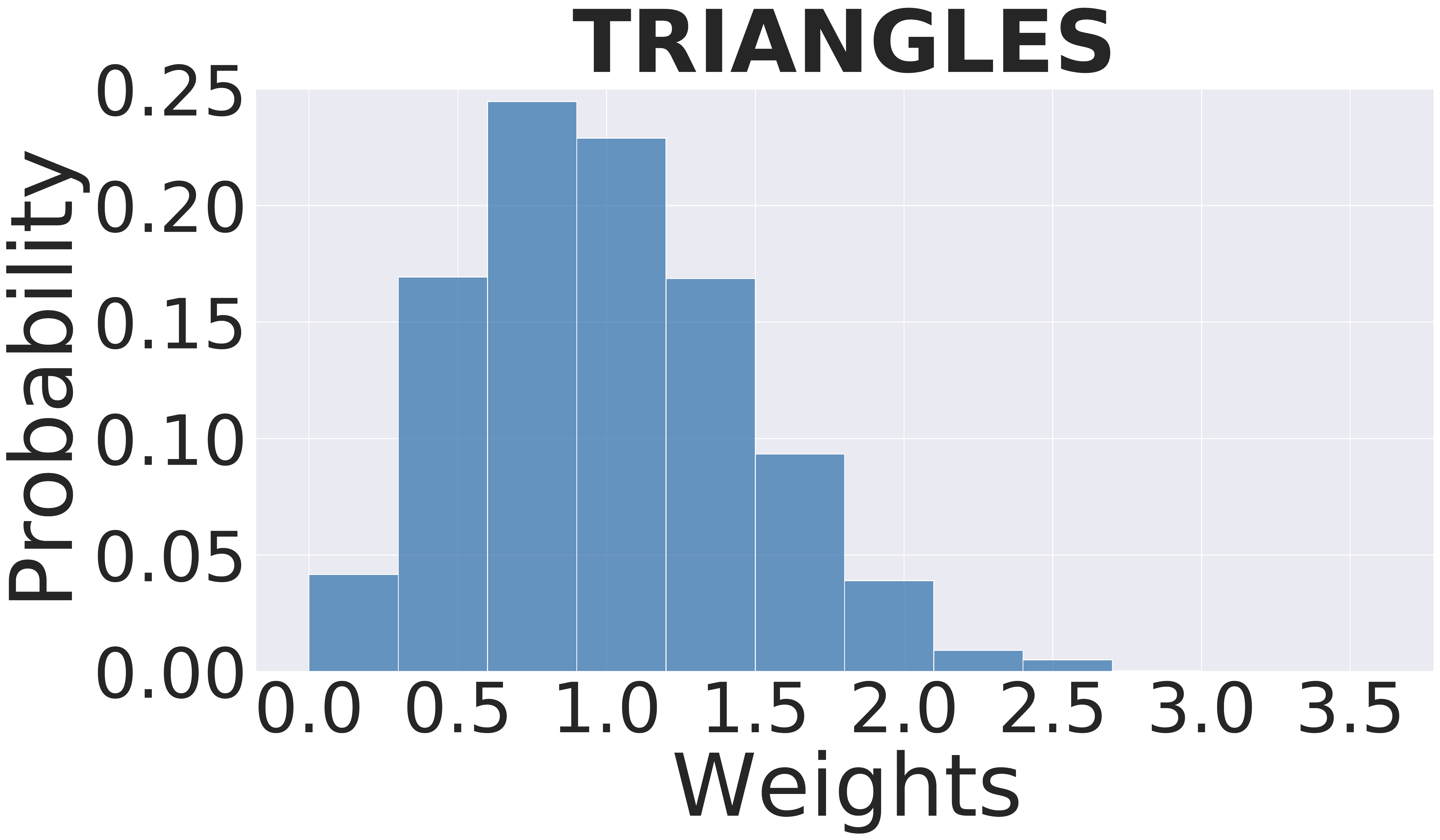}
        \caption{Weights distribution on TRIANGLES.}
    \end{subfigure}
    ~ 
    \begin{subfigure}[t]{0.32\textwidth}
        \centering
        \includegraphics[width=.99\textwidth]{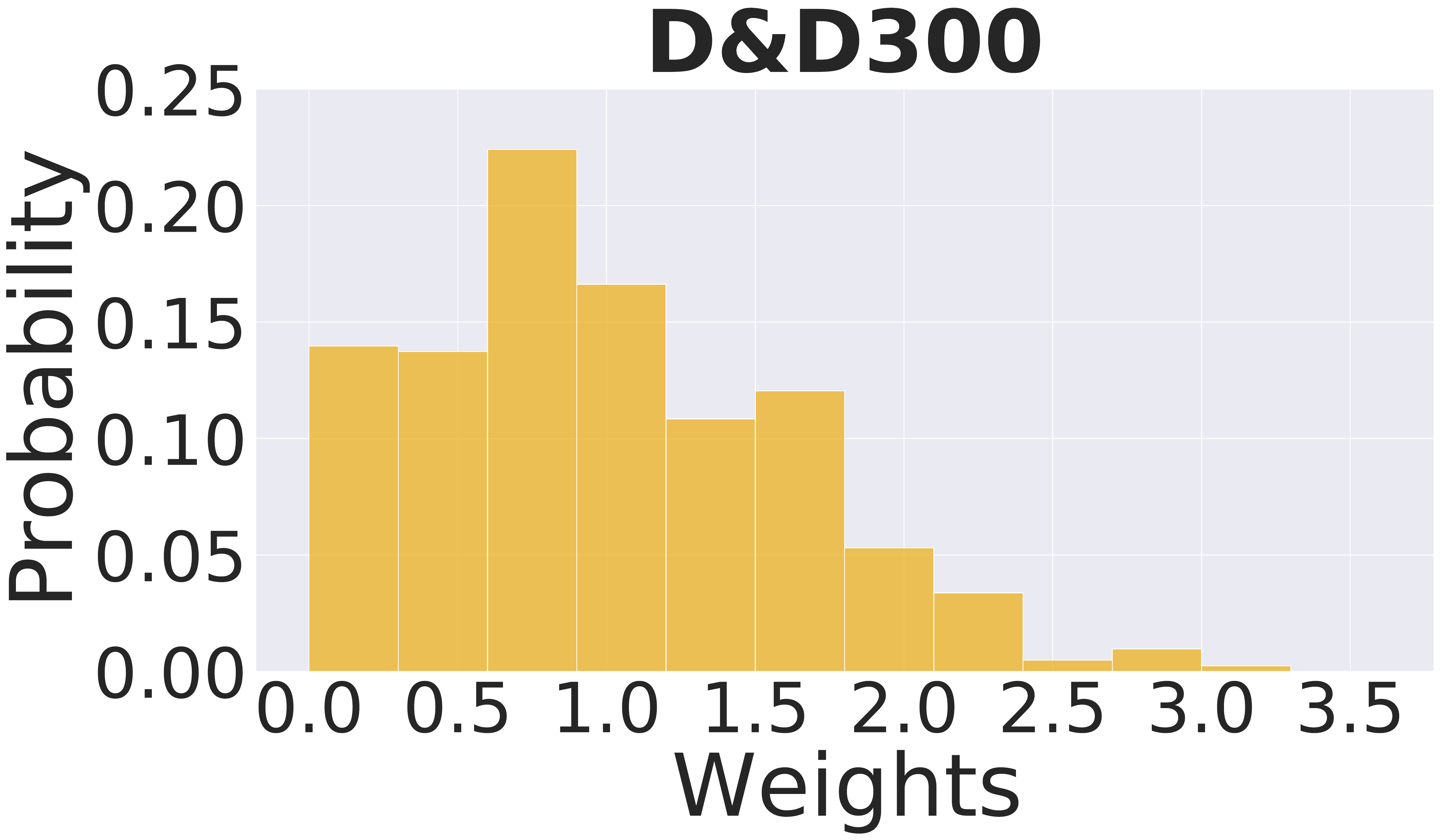}
        \caption{Weights distribution on D\&D$_{300}$.}
    \end{subfigure}
    ~ 
    \begin{subfigure}[t]{0.32\textwidth}
        \centering
        \includegraphics[width=.99\textwidth]{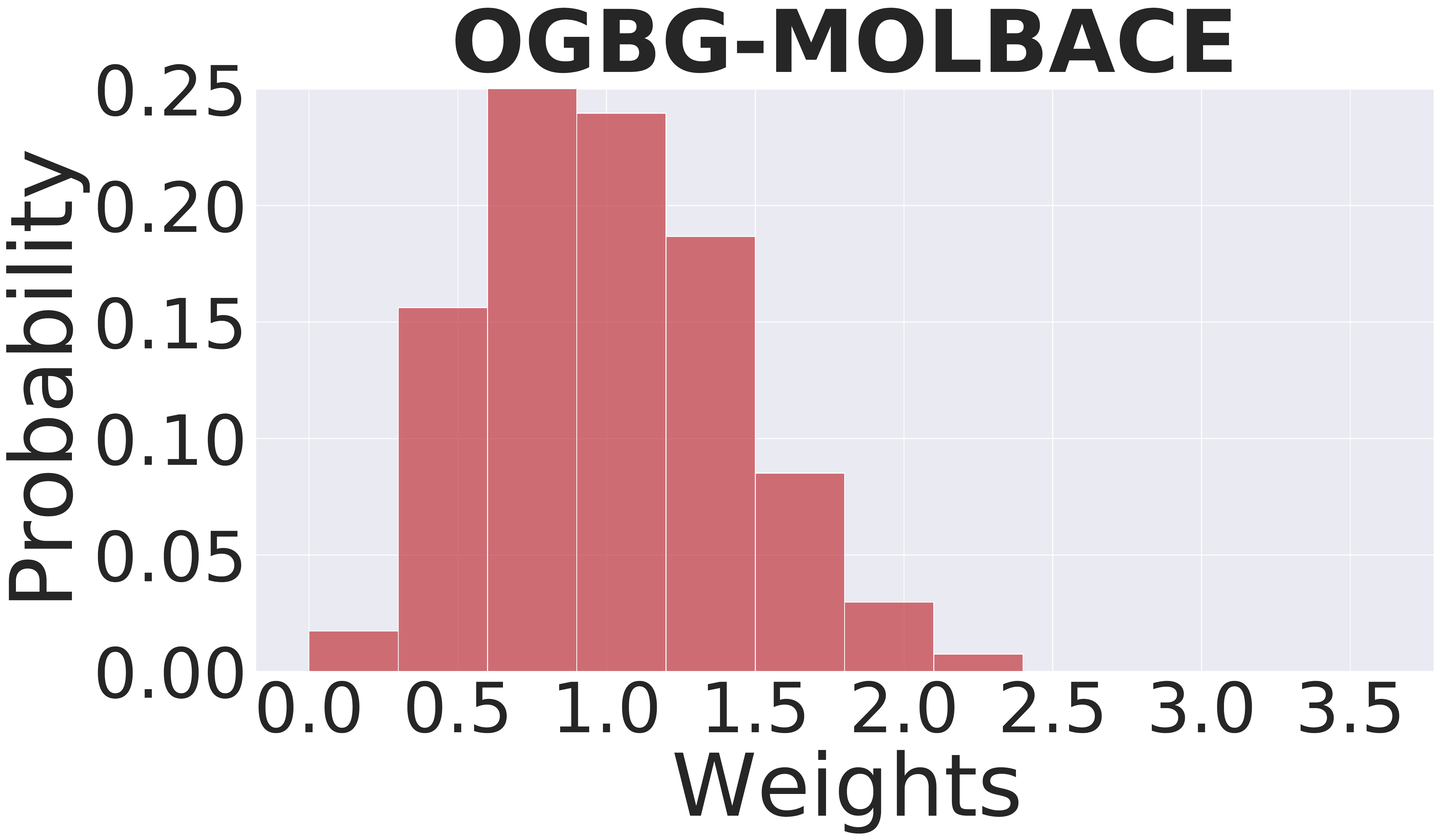}
        \caption{Weights distribution on OGBG-MOLBACE.}
    \end{subfigure}
\caption{The distribution of the learned graph weights after training on three datasets.}
\label{figure:weight_distribution}
\end{figure*}

\subsection{Weights Distribution}
In Figure~\ref{figure:weight_distribution} (a)(b)(c), to further investigate the effectiveness of the graph reweighting, we show the distribution of the learned graph weights on TRIANGLES, D\&D$_{300}$, and OGBG-MOLBACE when the training is finished. The results show that our proposed method learns non-trivial weights, and the weights distribution is slightly different across different datasets.

\subsection{Time Complexity}
Our method is not only effective but efficient to learn out-of-distribution generalized graph representation under complex distribution shifts.
The time complexity of our method is $O(\left|E\right|d+\left|V\right|d^2+K|\mathcal{B}|d^2)$, where $\left|V\right|$, $\left|E\right|$ denotes the total number of nodes and edges in the graphs, $d$ is the dimensionality of the representation, $K$ is the number of groups of global weights, and $|\mathcal{B}|$ is the batch size.
Specifically, the time complexity of the graph encoder GIN is $O(\left|E\right|d+\left|V\right|d^2)$ and the optimization of graph weights in Eq.~\eqref{equation:objective2} has $O(K|\mathcal{B}|d^2)$ complexity. 
As a comparison, the time complexity of GIN, our backbone GNN, is $O(\left|E\right|d+\left|V\right|d^2)$, i.e., our time complexity is on par since $d$, $K$, and $|\mathcal{B}|$ are small constants that are unrelated to the dataset size.

\subsection{Number of Parameters}

The parameters of our method consist of two parts, i.e., the graph encoder and graph weights. The former is determined by the graph encoder GNN architecture, which is GIN in our setting. The latter is determined by the number of graphs. Taking the OGBG-MOLBACE dataset for example, the number of parameters of our method is about 0.9M if we set the number of message-passing layers as 5 and the dimensionality of the representations as 300. 
Notice that our method has comparable or fewer parameters than the baselines. For the OGBG-MOLBACE dataset with the same hyper-parameter settings, GIN and PNA (two baselines in the experiments) have 0.9M and 6.0M parameters, respectively.
Nevertheless, our method achieves impressive out-of-distribution generalization performance against the baselines.

\subsection{Hyper-parameter Sensitivity}
\begin{figure*}[t!]
\centering
    \begin{subfigure}[t]{0.23\textwidth}
        \centering
        \includegraphics[width=.99\textwidth]{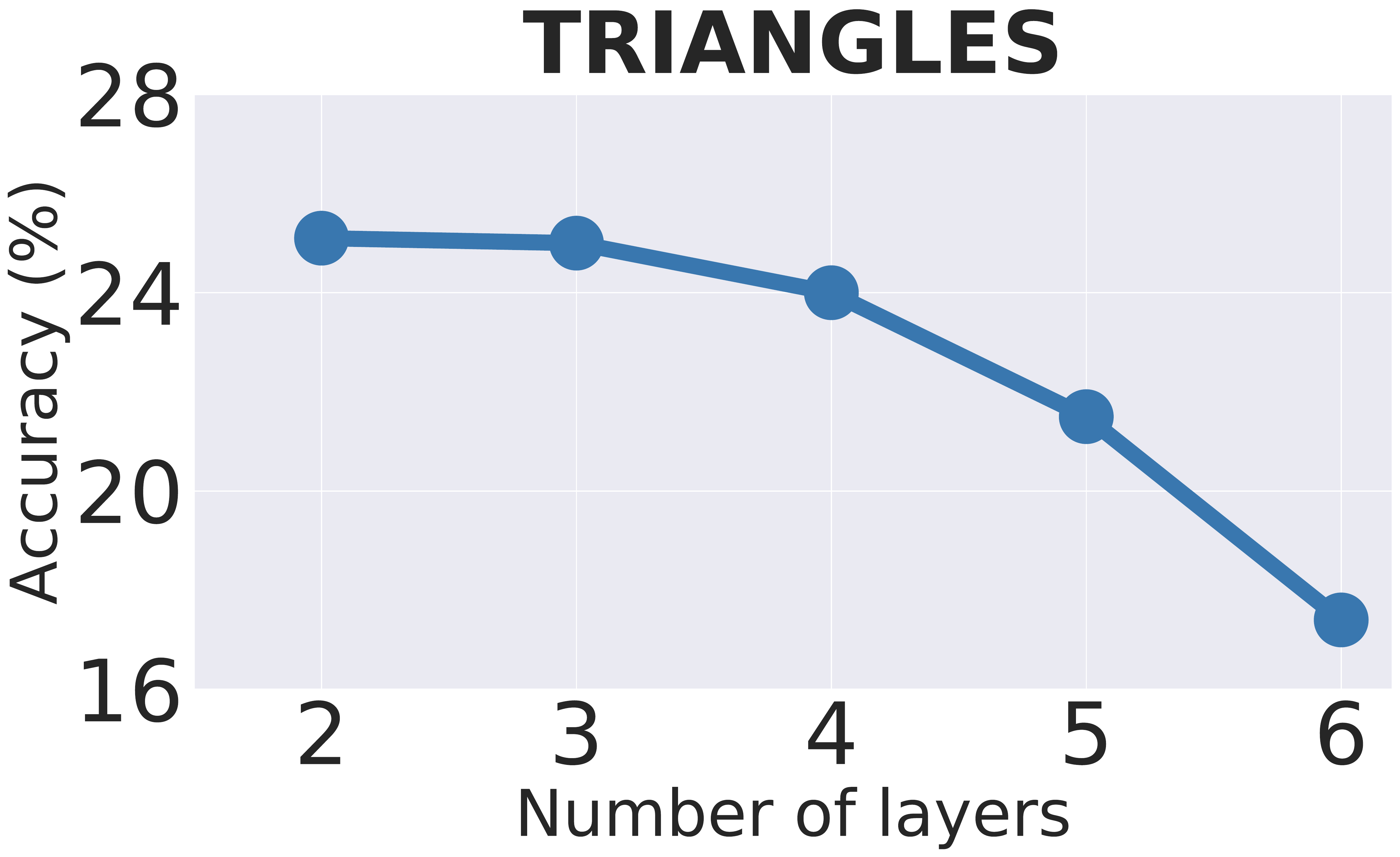}
        \caption{Number of layers.}
    \end{subfigure}
    ~ 
    \begin{subfigure}[t]{0.23\textwidth}
        \centering
        \includegraphics[width=.99\textwidth]{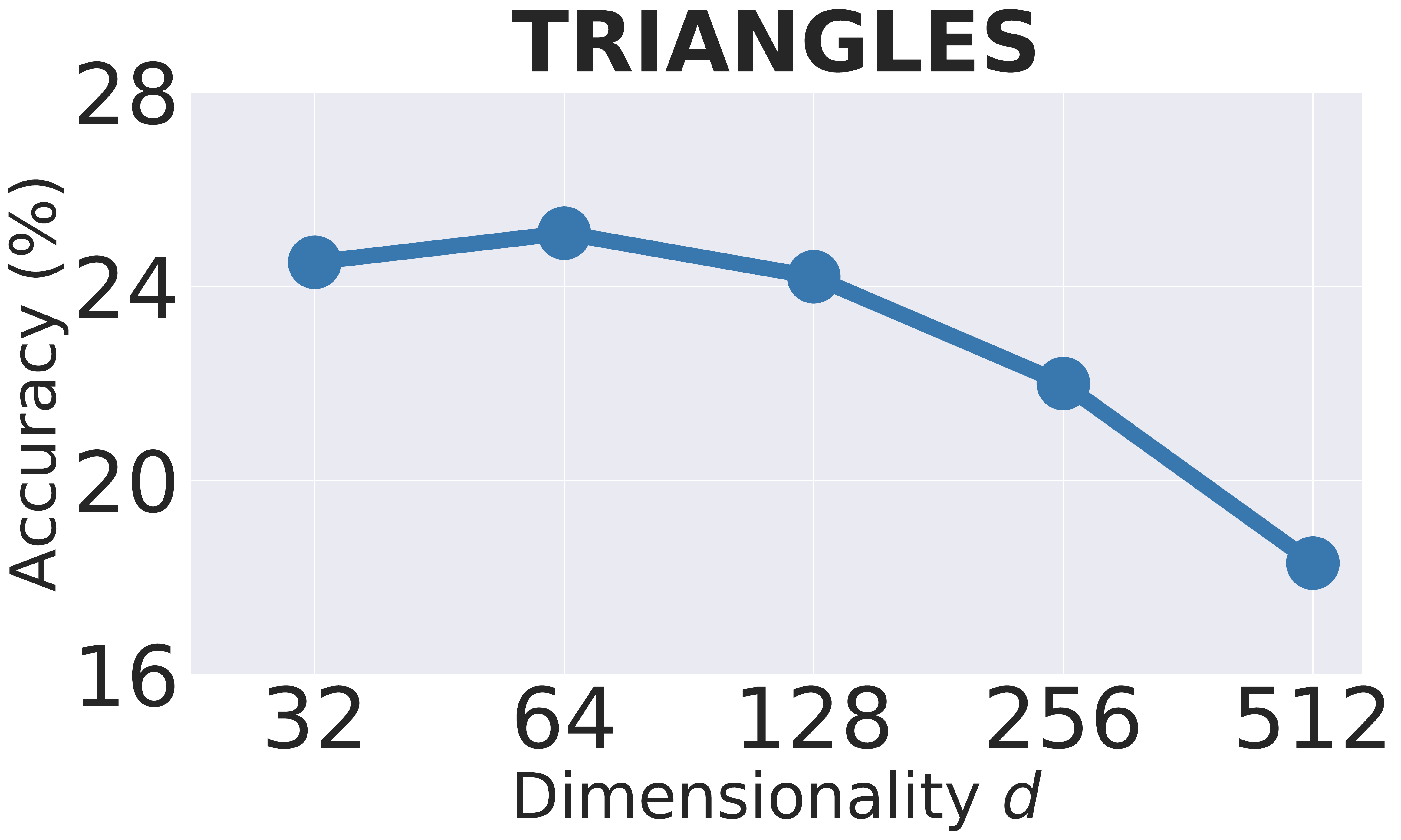}
        \caption{Dimensionality $d$.}
    \end{subfigure}
    ~ 
    \begin{subfigure}[t]{0.23\textwidth}
        \centering
        \includegraphics[width=.99\textwidth]{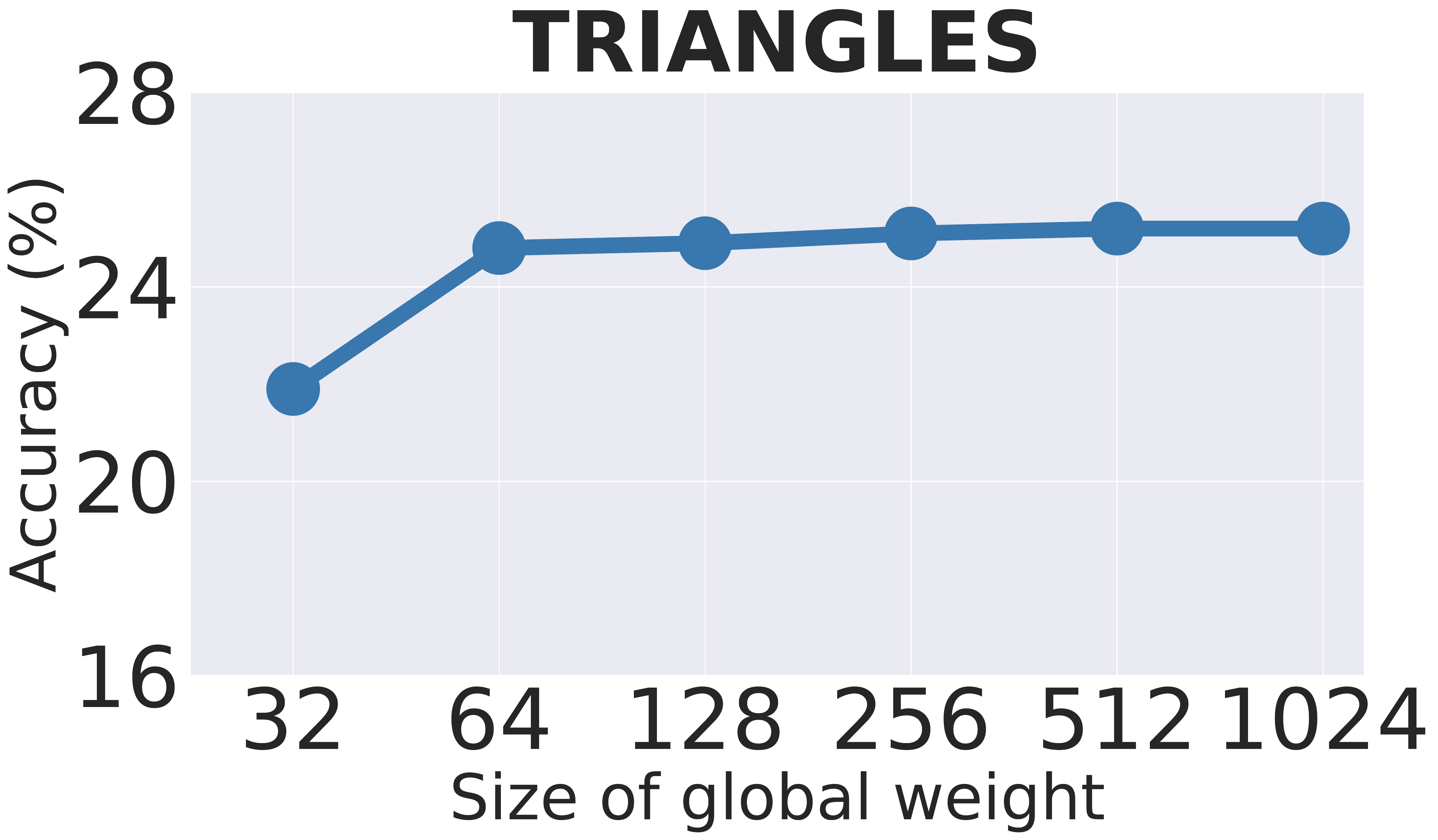}
        \caption{Size of global weights.}
    \end{subfigure}
    ~ 
    \begin{subfigure}[t]{0.23\textwidth}
        \centering
        \includegraphics[width=.99\textwidth]{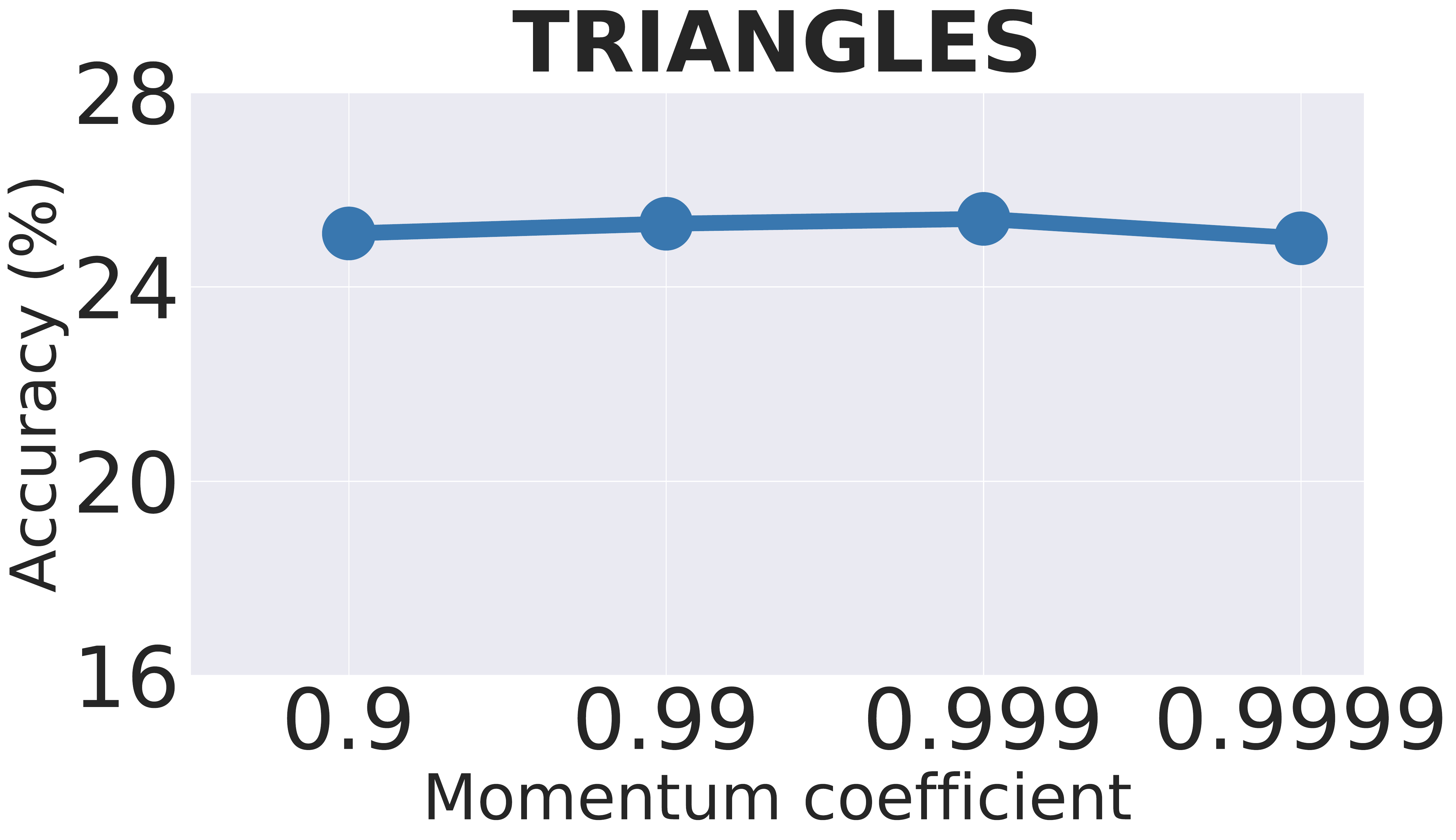}
        \caption{Momentum coefficient.}
    \end{subfigure}
\caption{The analyses of different hyper-parameters on TRIANGLES dataset.}
\label{figure:hypertriangles}
\end{figure*}

\begin{figure*}[t!]
\centering
    \begin{subfigure}[t]{0.23\textwidth}
        \centering
        \includegraphics[width=.99\textwidth]{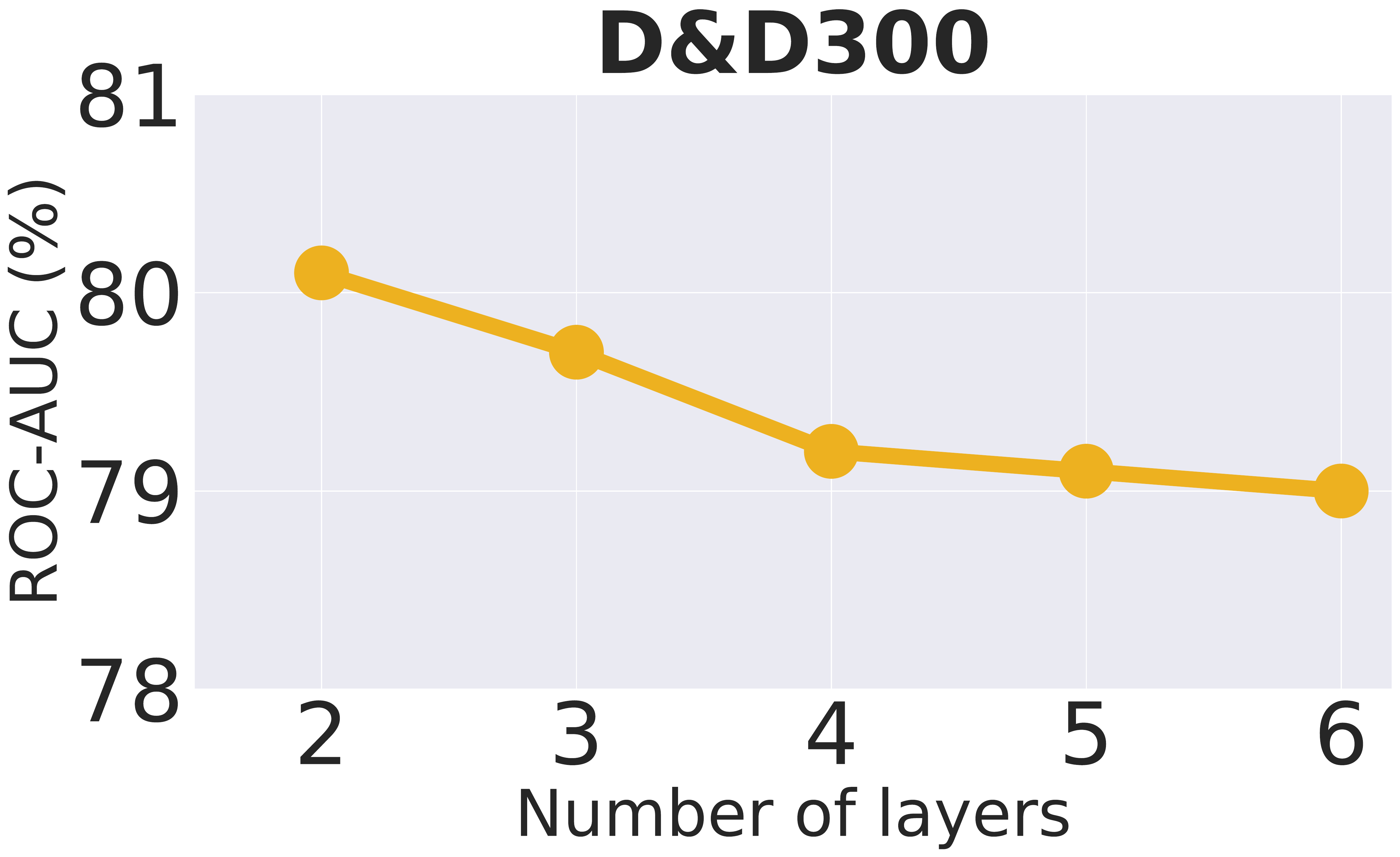}
        \caption{Number of layers.}
    \end{subfigure}
    ~ 
    \begin{subfigure}[t]{0.23\textwidth}
        \centering
        \includegraphics[width=.99\textwidth]{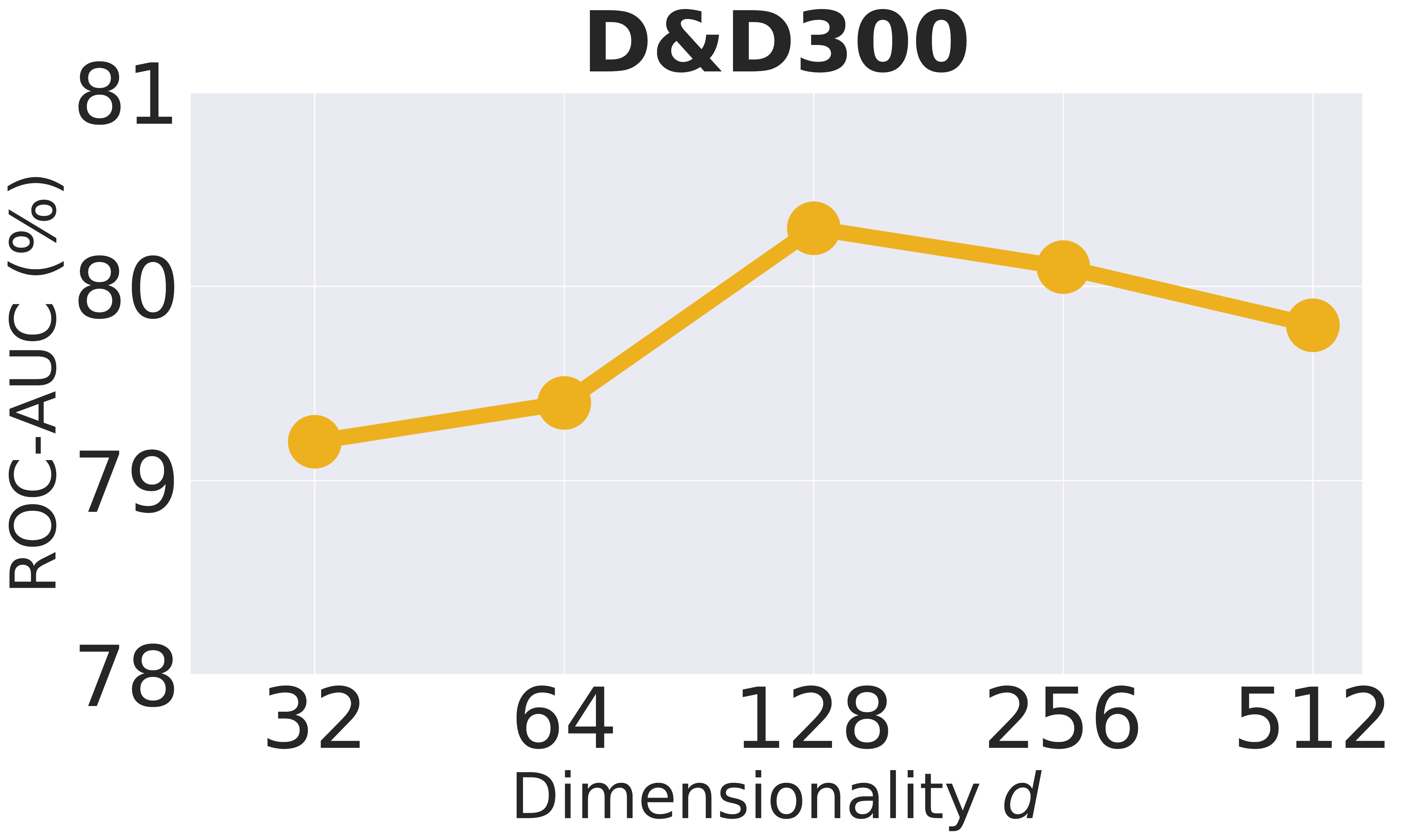}
        \caption{Dimensionality $d$.}
    \end{subfigure}
    ~ 
    \begin{subfigure}[t]{0.23\textwidth}
        \centering
        \includegraphics[width=.99\textwidth]{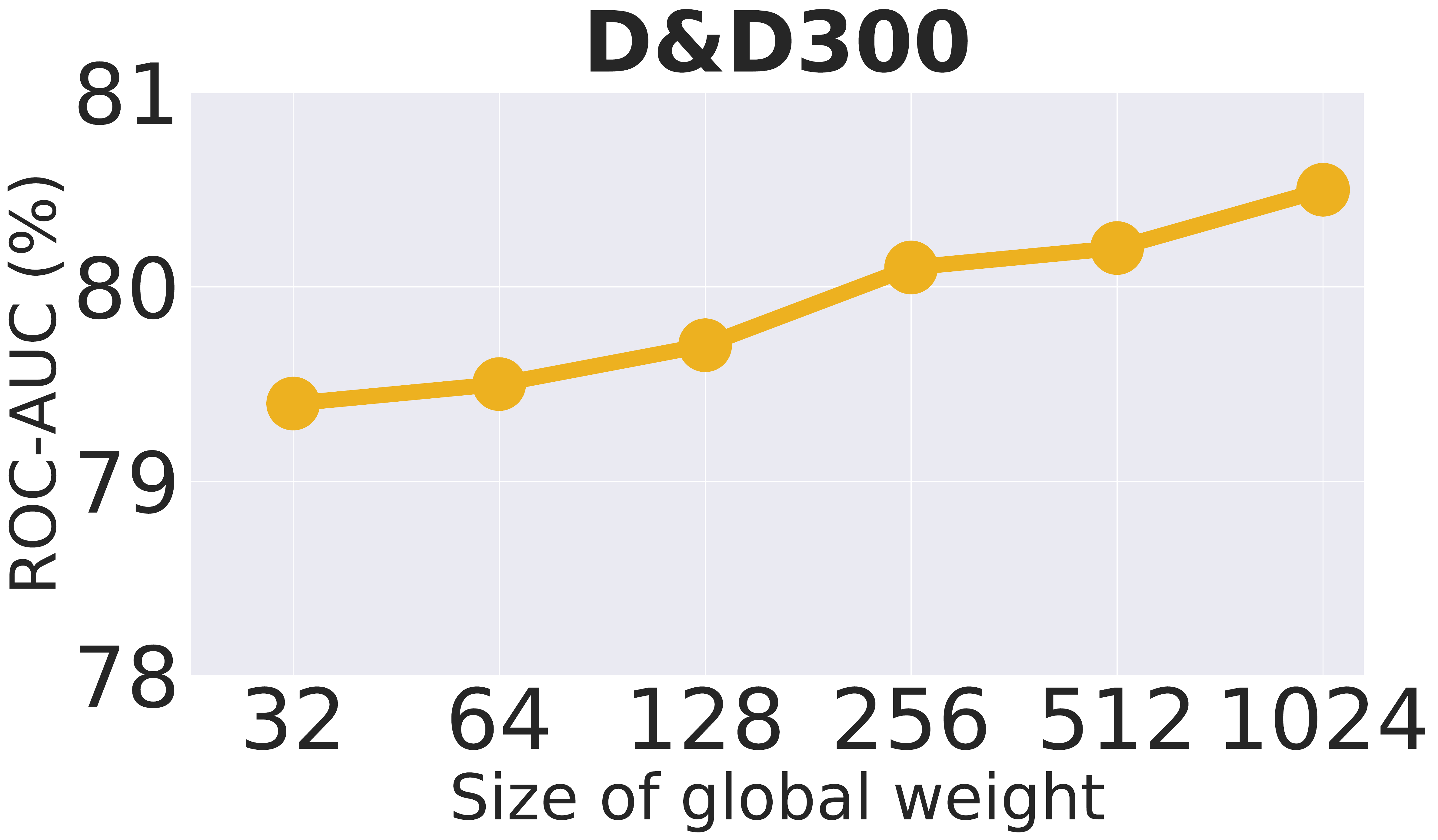}
        \caption{Size of global weights.}
    \end{subfigure}
    ~ 
    \begin{subfigure}[t]{0.23\textwidth}
        \centering
        \includegraphics[width=.99\textwidth]{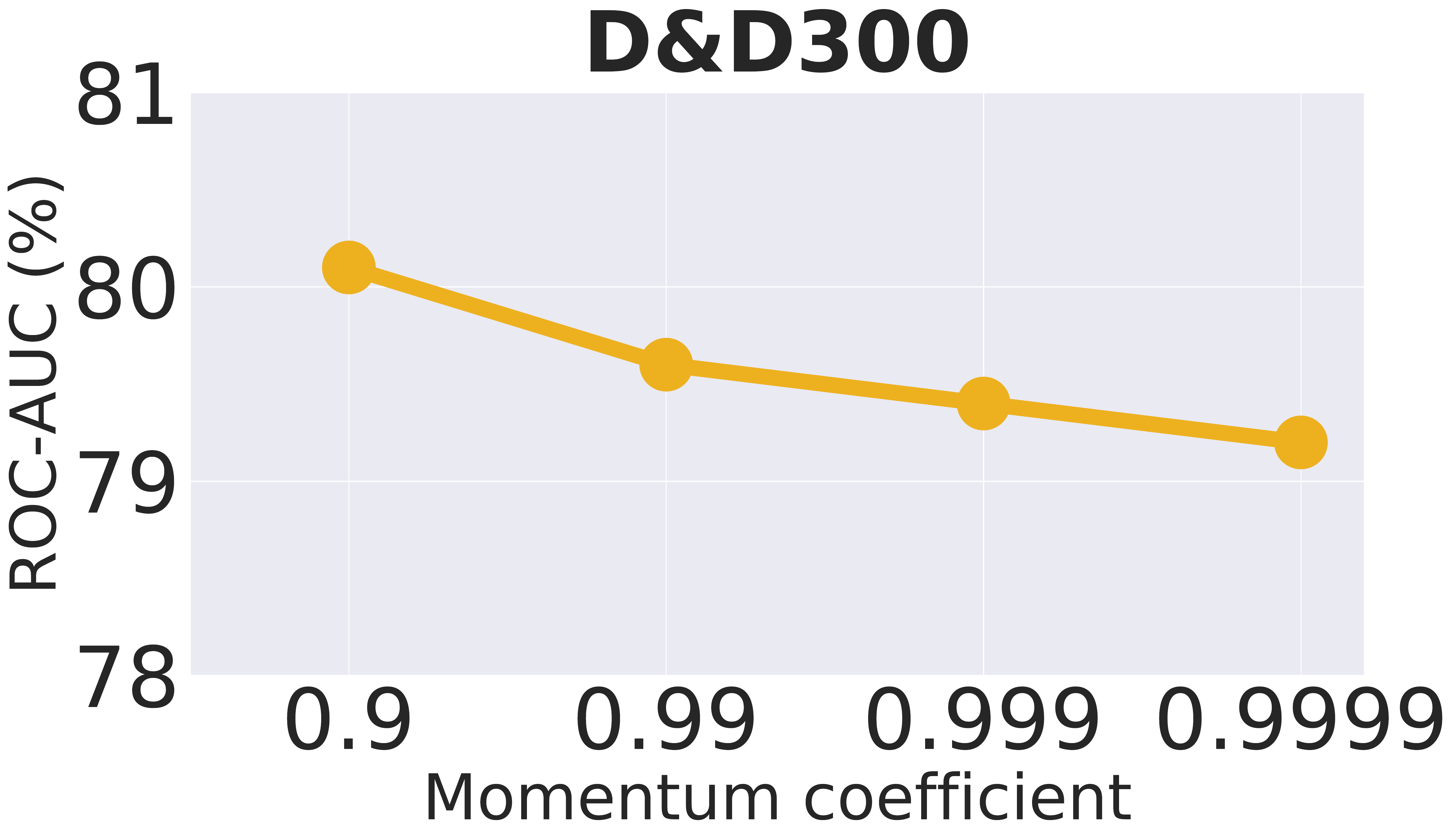}
        \caption{Momentum coefficient.}
    \end{subfigure}
\caption{The analyses of different hyper-parameters on D\&D$_{300}$ dataset.}
\label{figure:hyperdd}
\end{figure*}

\begin{figure*}[t!]
\centering
    \begin{subfigure}[t]{0.23\textwidth}
        \centering
        \includegraphics[width=.99\textwidth]{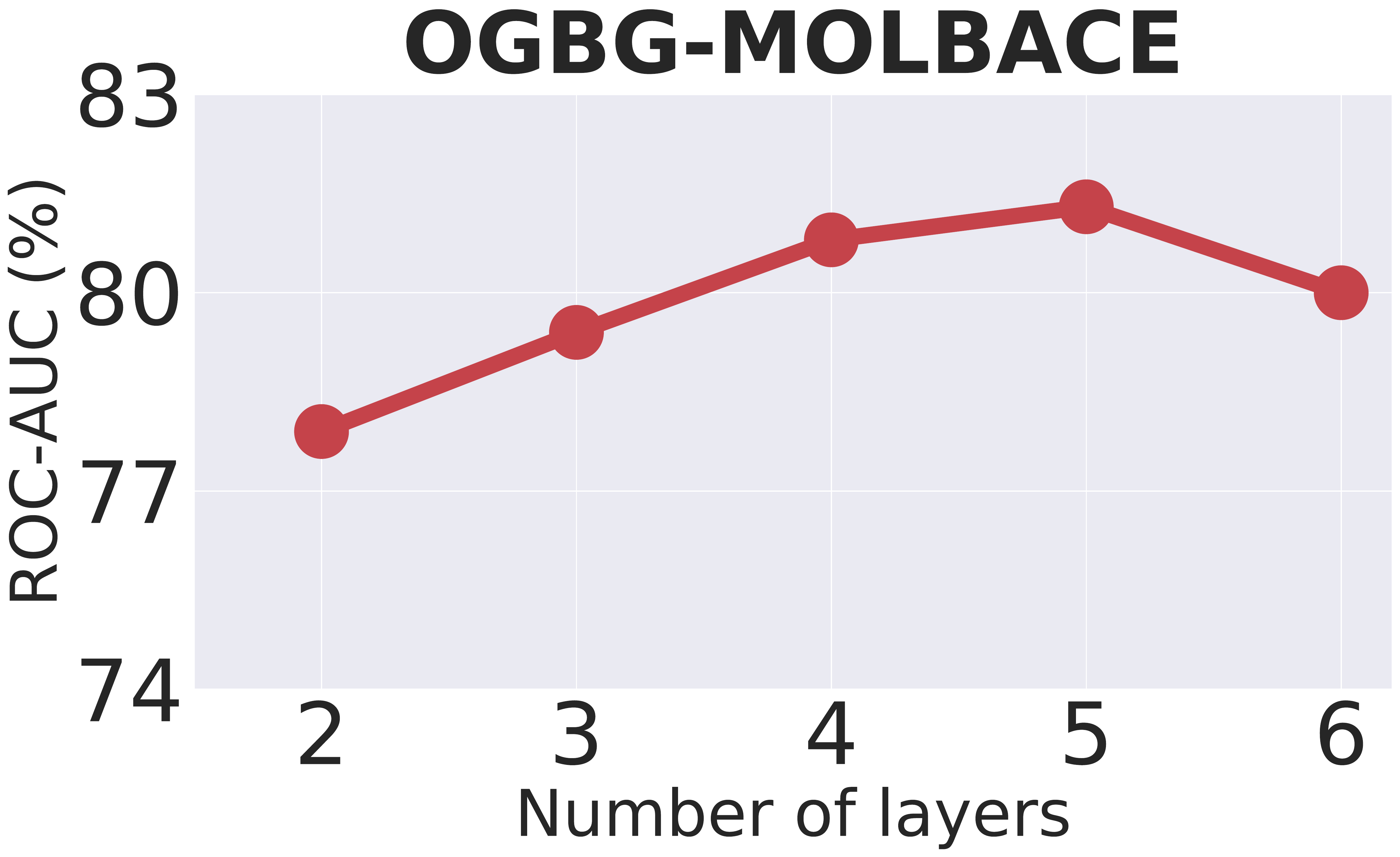}
        \caption{Number of layers.}
    \end{subfigure}
    ~ 
    \begin{subfigure}[t]{0.23\textwidth}
        \centering
        \includegraphics[width=.99\textwidth]{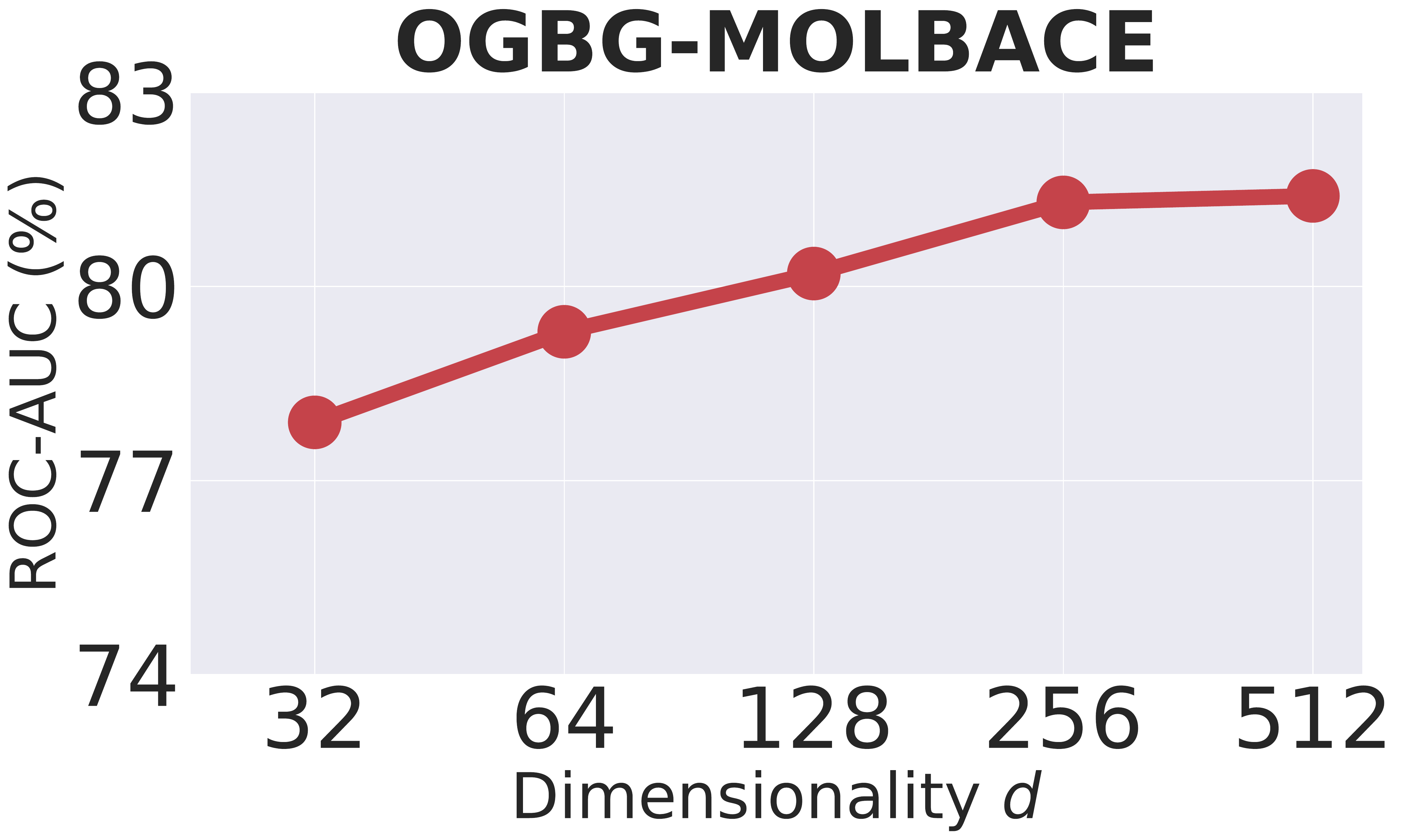}
        \caption{Dimensionality $d$.}
    \end{subfigure}
    ~ 
    \begin{subfigure}[t]{0.23\textwidth}
        \centering
        \includegraphics[width=.99\textwidth]{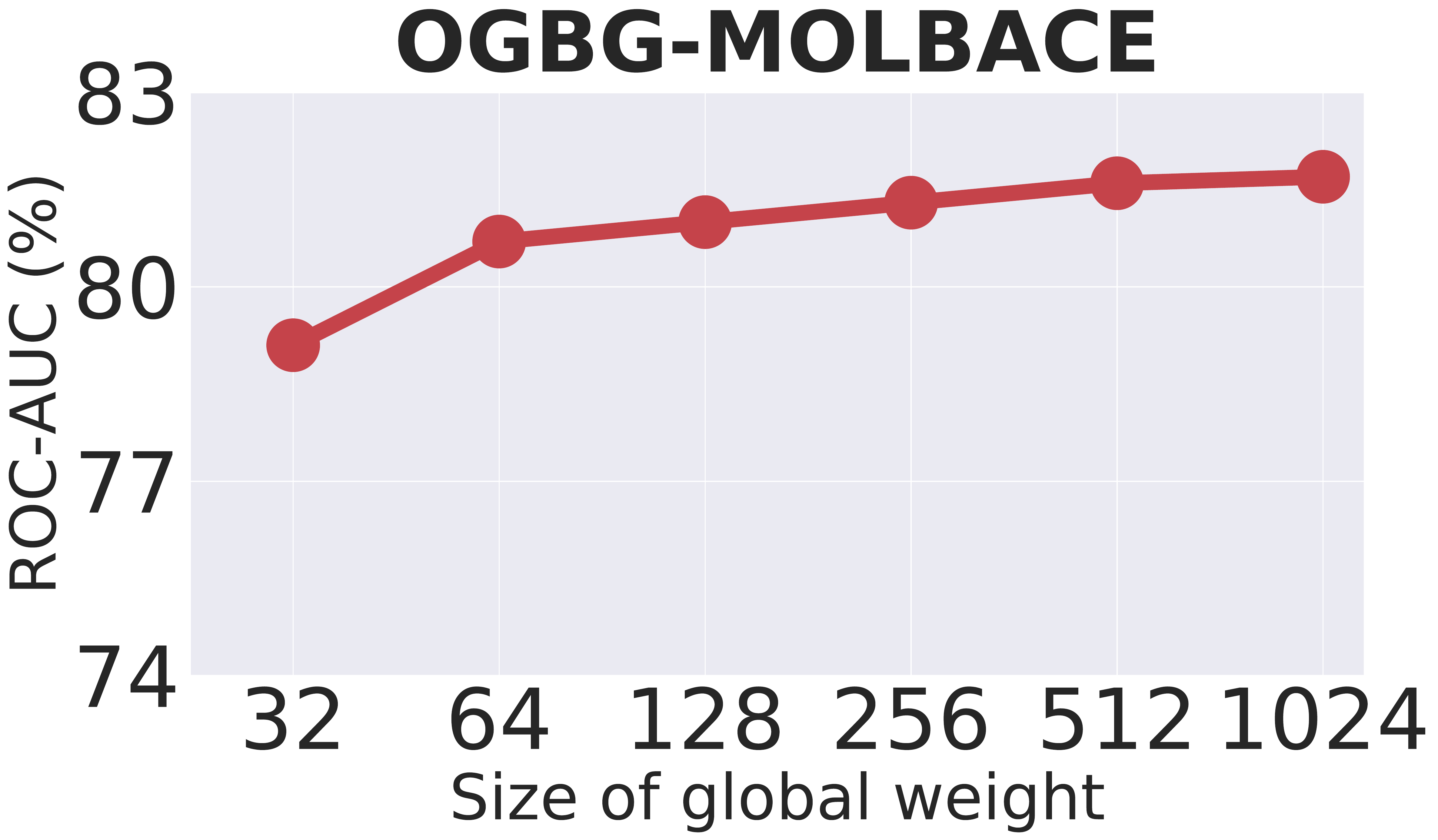}
        \caption{Size of global weights.}
    \end{subfigure}
    ~ 
    \begin{subfigure}[t]{0.23\textwidth}
        \centering
        \includegraphics[width=.99\textwidth]{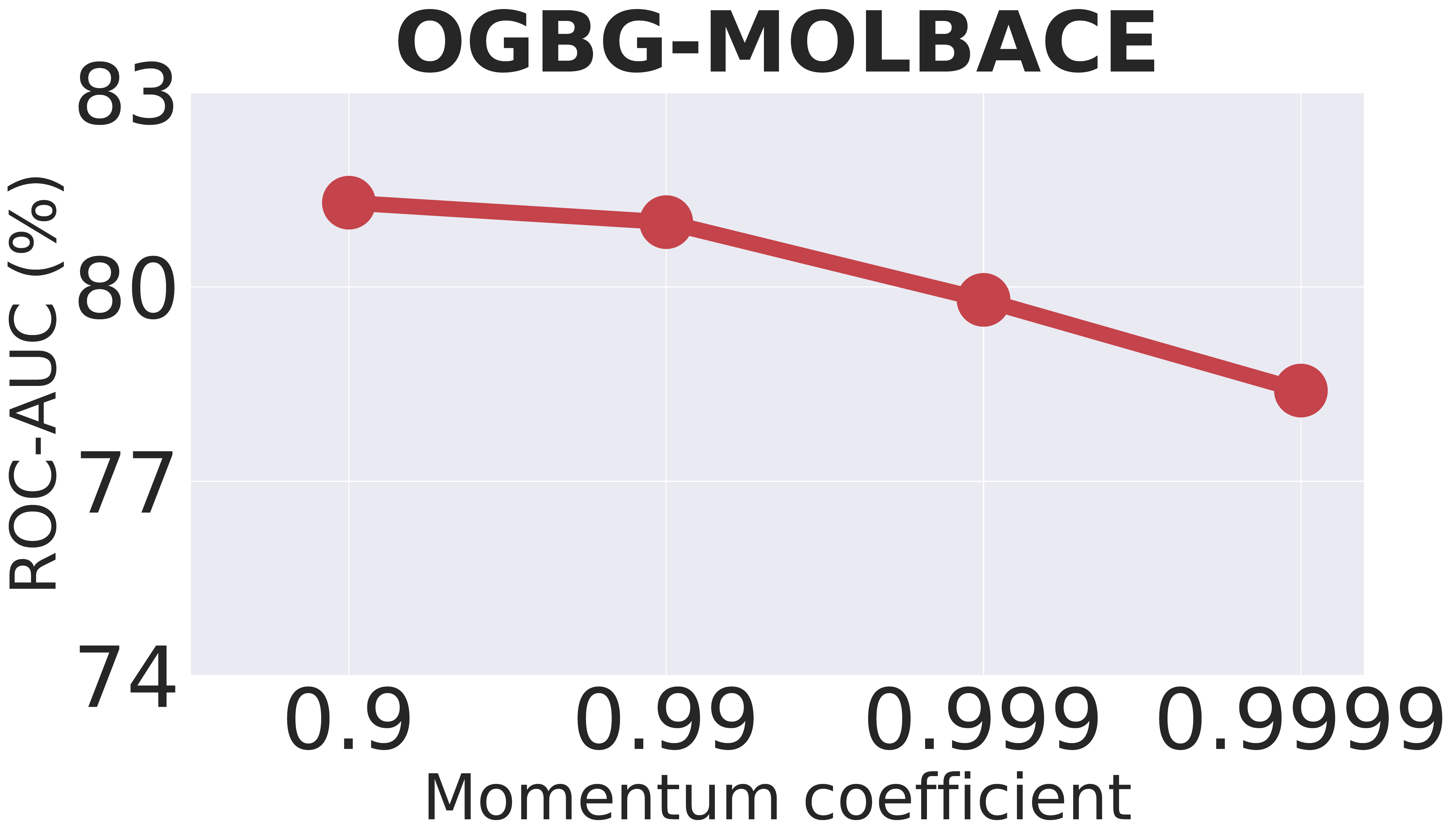}
        \caption{Momentum coefficient.}
    \end{subfigure}
\caption{The analyses of different hyper-parameters on OGBG-MOLBACE dataset.}
\label{figure:hyperogb}
\end{figure*}

We investigate the sensitivity of hyper-parameters of our method, including the number of message-passing layers in the graph encoder, the dimensionality of the representations $d$, the size of global weights, and the momentum coefficient $\gamma$ in updating global weights. 
For simplicity, we only report the results on TRIANGLES (see Figure~\ref{figure:hypertriangles}), D\&D$_{300}$ (see Figure~\ref{figure:hyperdd}), and OGBG-MOLBACE (see Figure \ref{figure:hyperogb}), while the results on other datasets show similar patterns.
From Figures \ref{figure:hypertriangles}--\ref{figure:hyperogb}, we observe that the performance relies on an appropriate choice of the number of message-passing layers of the graph encoder. Since the task of counting triangles is relatively simple, the graph encoder with two message-passing layers is good enough on TRIANGLES, while five layers are needed to achieve the best performance on OGBG-MOLBACE.
When the number of layers of graph encoder is small, the model has limited capacity and may not be able to fuse enough information from neighbors. On the other hand, a very large number of layers could lead to the over-smoothing problem~\cite{miao2021lasagne}.
Besides, the optimal dimensionality of the representations $d$ for TRIANGLES is relatively smaller than that for D\&D$_{300}$ and OGBG-MOLBACE. 

In addition, as the size of global weights increases, the performance is improved.
The global representations and weights can help to learn consistent graph sample weights on the whole dataset and therefore improve the generalization ability of the model.

Finally, we find that the momentum coefficient $\gamma$ also has a slight influence on the performance. A large $\gamma$ will make the update of global representations and weights slower, and a small one will accelerate the update, corresponding to emphasizing long-term and short-term memory, respectively.

\section{Conclusions}
\label{section:conclusions}
In this paper, we propose a novel out-of-distribution generalized graph neural network (\mymodel) to solve the problem of generalization of GNNs under complex and heterogeneous distribution shifts.
We propose a nonlinear graph representation decorrelation method by utilizing random Fourier features and sample reweighting, so that the learned representations of \model are encouraged to eliminate the statistical dependence between the representations.
We further present a scalable global-local weight estimator, which can learn graph weights for the whole dataset consistently and efficiently.
Extensive experiments on both synthetic and real-world datasets demonstrate the superiority of our method against state-of-the-art baselines for out-of-distribution generalization.

 \section*{Acknowledgments}
We would like to thank Xingxuan Zhang for valuable discussions.
This work was supported in part by the National Key Research and Development Program of China No. 2020AAA0106300 and National Natural Science Foundation of China (No. 62050110, No. 62102222).
All opinions, findings, conclusions and recommendations in this paper are those of the authors and do not necessarily reflect the views of the funding agencies.

\medskip
{
\bibliographystyle{unsrtnat}
\bibliography{Reference}
}

\end{document}